\documentclass[twocolumn]{article}
\usepackage[T1]{fontenc}
\usepackage{lmodern}
\usepackage[square,sort,comma,numbers]{natbib}

\usepackage{balance}
\usepackage[utf8]{inputenc} 
\usepackage[T1]{fontenc}    
\usepackage{hyperref}       
\usepackage{url}            
\usepackage{booktabs}       
\usepackage{amsfonts}       
\usepackage{nicefrac}       
\usepackage{microtype}      
\usepackage{xcolor}         

\usepackage{amstext}
\usepackage{amsmath}
\usepackage{graphicx}
\usepackage{paralist}
\newcommand{\vir}[1]{``#1''}

\newcommand{\bSigma}{\boldsymbol{\Sigma}}
\newcommand{\bI}{\boldsymbol{I}}

\newcommand{\bx}{\boldsymbol{x}}

\newcommand{\bv}{\textit{\textbf{v}}}
\long\def\ignore#1{}

\usepackage{subfigure}

\usepackage{xr}
\makeatletter
\newcommand*{\addFileDependency}[1]{
  \typeout{(#1)}
  \@addtofilelist{#1}
  \IfFileExists{#1}{}{\typeout{No file #1.}}
}
\makeatother

\begin{document}

\title{A Critical Analysis of Classifier Selection in Learned Bloom Filters}

\author{Dario Malchiodi$^{1,3}$, Davide Raimondi$^1$, Giacomo Fumagalli$^1$,
Raffaele Giancarlo$^2$, Marco Frasca$^1$ \\
$^1$ Dept. of Computer Science, University of Milan, Milano, Italy \\
$^2$ Dept. of Mathematics and CS, University of Palermo, Palermo, Italy \\
$^3$ DSRC, University of Milan, Milano, Italy \\
$\{$dario.malchiodi, marco.frasca$\}$@unimi.it, \\
$\{$davide.raimondi2, giacomo.fumagalli1$\}$@studenti.unimi.it, \\
raffaele.giancarlo@unipa.it}

\date{}

\maketitle

\begin{abstract}
Learned Bloom Filters, i.e.,  models induced from data via machine learning techniques and solving the approximate set membership problem, have recently been introduced with the aim of enhancing the performance of standard Bloom Filters, with special focus on space occupancy. Unlike in the classical case, the \vir{complexity} of the data used to build the filter might heavily impact on its performance. Therefore, here  we propose the first in-depth analysis, to the best of our knowledge, for the performance assessment of a given Learned Bloom Filter, in conjunction with a given classifier, on a dataset of a given classification complexity. Indeed, we propose a novel methodology, supported by software, for designing, analyzing and implementing Learned Bloom Filters in function of specific constraints on their multi-criteria nature (that is, constraints involving space efficiency, false positive rate, and reject time). Our experiments show that the proposed methodology  and the supporting software are valid and useful:  we find out that only two classifiers have desirable properties in relation to problems with different data complexity, and, interestingly, none of them has been considered so far in the literature. We also experimentally show that the Sandwiched variant of Learned Bloom filters is the most robust to data complexity and classifier performance variability, as well as those usually having smaller reject times. The software can be readily used to test new Learned Bloom Filter proposals, which can be compared with the best ones identified here.
\end{abstract}

\section{Introduction}\label{sec:intro}
Recent studies have highlighted how the impact of machine learning has the potential to change the way we design and analyze algorithms and data structures. Indeed, the resulting research area of Learned Data Structures has had a well documented impact on a broad and strategic domain such as that of Data Bases. An analogous impact  can be expected for Network Management \cite{wu21} and Computational Biology \cite{btaa911}.
More in general, as well argued in \cite{Kraska21}, this novel way to use machine learning has the potential to change how Data Systems are designed.
This new trend  was initiated by Kraska et. al. \cite{Kraska18}, as far as Data Structures are concerned, and then extended to Algorithms by Mitzenmacher and S. Vassilvitskii \cite{Mitz20}. Concerning the former, the common theme to this new approach is that of training a  Classifier \cite{duda20} or a Regression Model~\cite{FreedmanStat} on the input data. Then such a learned model is used as an \vir{oracle} that a given  \vir{classical} data structure can use  in order to answer queries with improved performance (usually time). To date, Learned Indexes have been the most studied, e.g.,  \cite{amato2021lncs,Ferragina:2020pgm,FERRAGINA21,Kipf20,Kirshe20, Kraska18,Mailtry21,Marcus20,Markus20b}, although Rank/select data structures have also received some attention \cite{Boffa21}. In this work, we focus on Bloom Filters (BFs) \cite{Bloom70} which, as detailed in what follows, have also received attention in the realm of Learned Data Structures. Such an attention is quite natural, due to the fundamental nature and pervasive use of Bloom Filters. Indeed,  many variants and alternatives for these filters have been already proposed, prior to the Learned versions \cite{Broder2005}.

{\bf Problem Statement, Performance of a Bloom Filter and a Learned Version.}
Bloom Filters (BF) solve the \emph{Approximate Set Membership} problem, defined as follows: having fixed a \emph{universe} $U$ and a set of \textit{keys} $S \subset U$, for any given $\bx \in U$, find out whether or not $\bx \in S$.  \emph{False negatives}, that is negative answers when $\bx \in S$, are not allowed. On the other hand, \emph{false positives} (i.e., elements in $U \backslash S$ wrongly decreed as keys) are allowed, albeit their fraction (termed henceforth \emph{false positive rate}, FPR for short) should be bounded by a given $\epsilon$.
The key parameters of any data structure solving the approximate set membership problem are:
\begin{inparaenum}[(i)]
\item the FPR $\epsilon$;
\item the total space needed by the data structure; and
\item the \emph{reject time}, defined as the expected time needed to reject a non-member of $S$.
\end{inparaenum}


Kraska et al. \cite{Kraska18} have proposed a Learned version of Bloom Filters (LBF) in which a suitably trained  binary classifier is introduced with the aim of reducing space occupancy w.r.t.\ a classical BF, having fixed the FPR.
Such classifier is initially queried to predict the set membership, with a fallback to a standard BF in order to avoid false negatives. Mitzenmacher \cite{Mitz18} has provided a model for those filters, together with a very informative mathematical analysis of their pros/cons, resulting in new models for LBFs, and additional models have been introduced  recently \cite{Dai,Kraskap}. It is worth pointing out that all the mentioned Learned Bloom Filters are static, i.e., no updates are allowed. This is the realm we are studying here, aiming at a suitable joint optimization of the aforementioned key resources. As for the dynamic case, some progress is reported in \cite{Liu22} in the Data Stream model of computation.

{\bf The Central Role of Classifier Selection.}
It is worth pointing out that, although they differ in architecture, each of these proposals has a binary classifier at its core. Somehow, not much attention has been devoted to the choice of the classifier to be used in practical settings, despite its centrality in this new family of filters and its role in the related theoretical analysis \cite{Mitz18}. Kraska et al.\ use  a Recurrent Neural Network, while Dai and Shrivastava \cite{Dai} and Vaidya et al.~\cite{Kraskap} use Random Forests. These choices are only informally motivated, giving no evidence of superiority with respect to other possible ones, e.g., via a comparative analysis.
Therefore, apart from an initial study presented in \cite{Fumagalli:2021}, the important problem of suitably choosing the classifier to be used to build a specific LBF has not been fully addressed so far. In addition to that, although the entire area of Learned Data Structures and Algorithms finds its methodological  motivation as a conceptual tool to reconsider classic approaches in a data-driven way, the role that the \emph{complexity} of a dataset plays in guiding the practical choice of a Learned Data Structure
for that dataset has been considered to some extent for Learned Indexes only \cite{Mailtry21}.
This aspect is even more relevant for LBFs. Indeed, as well discussed in \cite{Mitz18}, while the performance of classic BFs is  \vir{agnostic} w.t.r.\ the statistical properties of the input data, LBFs are quite dependent on them. In addition, it is well known that the performance of a learnt classifier (a central component in this context) is very sensitive to the \vir{classification complexity} of a dataset~\cite{ali2006,Cano13,flores2014,luengo2015}.
Such a State of the Art is problematic, both methodologically and practically, for LBFs to be a reliable competitor of their classic counterparts.

\noindent \textbf{Goals:}
Our aim is to provide a methodology and the associated software to support the design, analysis and deployment of Learned Boom Filters with respect to specific constraints regarding their multi-criteria nature.  Namely,  space efficiency,  false positive rate, and reject time.

\noindent \textbf{Contributions:} In order to achieve these goals, our contributions are the following.
\begin{enumerate}[(1)]
    \item \textbf{We revisit Bloom Filters}, both in their original and learned versions (Section~\ref{sec:BF}), detailing the hyperparameters to be tuned within the related training procedures.
    \item \textbf{We propose a methodology}, which can guide both designers and users of LBFs in their design choices (Section~\ref{sec:methodology}), to study the interplay among:
 \begin{enumerate}[(a)]
 \item the parameters indicating  how a filter, learned or classic, performs on an input dataset;
 \item the classifier used to build the LBF;
 \item the classification complexity of the dataset.
 \end{enumerate}
     \item \textbf{Software platform and findings}: we provide a software platform implementing the above-mentioned methodology and we show the validity and usefulness of our approach, as detailed next.
     \begin{enumerate}
        \item We address the problem of choosing the most appropriate classifier in order to design a LBF having as only prior knowledge the total space budget, the data complexity and the list of available classifiers, including as selection criterion  also the reject time. A related problem has been considered in~\cite{Mitz18} with two important differences: the filter is fixed, and the obtained results supply only partial answers, leading to the suggestion of an experimental methodology, which has not been validated and it is not supported by software.  Our analysis shows the following.
    \begin{itemize}

 \item Among the many classifiers used in this research,  only two classifiers are worth of attention. Remarkably, none of the two  has been considered before for Learned Bloom Filters (Section~\ref{sec:complex}).

 \end{itemize}

    \item As a further contribution, we assess how the performance of State-of-the-Art BFs is affected by datasets of increasing complexity (Section~\ref{sec:complex}).
   \begin{itemize}
\item We identify a version  of Learned Bloom Filter more robust  than existing proposals to variations of data complexity and classifier performance.
   \end{itemize}

    \end{enumerate}
\end{enumerate}

In conclusion, our experiments show that the proposed methodology is valid for the design of Learned Bloom Filters and that the associated software is quite useful. Indeed, in addition to the novel findings mentioned above, we can also  provide  recommendations on how to use State-of-the-Art solutions (Section~\ref{sec:filters_eval}).

\section{Bloom Filters and Learned Bloom filters}\label{sec:BF}
\subsection{Bloom Filters}
A Bloom Filter~\cite{Bloom70} is a data structure solving the Approximate Set Membership problem defined in the Introduction, based on a boolean array $\bv$ of $m$ entries and on $k$ hash functions $h_1, \dots, h_k$ mapping $U$ to $\{1, \dots, m\}$. These functions are usually assumed to be \textit{k-wise independent}~\cite{wegman81,wegman79}, although much less demanding schemes work well in practice~\cite{Bloom70}.
A BF is built by initializing all the entries of $\bv$ to  zero, subsequently considering all keys $\bx \in S$ and setting $v_{h_j(\bx)} \leftarrow 1$ for each $j \in \{1, \dots k\}$; a location can be set to $1$ several
times, but only the first change has an effect. Once the filter has been built, any $\bx \in U$ is tested against membership in $S$ by evaluating the entry $v_{h_j(\bx)}$, for each hash function $h_j$: $\bx$ is classified as a key if all tested entries are equal to $1$, and rejected (a shorthand for saying that it is classified as a non-key) otherwise. False positives might arise because of hash collisions, and the corresponding rate $\epsilon$  is inversely bound to the array size $m$. More precisely, equation (21) in~\cite{Bloom70} connects
reject time, space occupancy and FPR, so that one can choose the configuration of the filter: for instance, given the available space, one can derive the reject time that minimizes the FPR. Analogous trade-offs~\cite {Broder2005,Mitz18} can be used to tune the hyperparameters of a BF (namely, $m$ and $k$) in order to drive the inference process towards the most space-conscious solution. In particular, fixed an FPR $\epsilon$ and a number $n = |S|$ of keys, a BF ensuring optimal reject time requires an array of
\begin{equation}\label{eq:bloom}
    m = 1.44n\log(1/\epsilon) \text{ bits.}
\end{equation}

\subsection{Learned Bloom Filters}\label{sec:LBF}
A Learned Bloom Filter~\cite{Kraska18} is a data structure simulating a BF to reduce its resource demand or its FPR by leveraging a classifier.
The main components of a LBF are a classifier $C: U \to [0, 1]$ and a BF $F$, defined as follows.
\begin{enumerate}
\item Using supervised machine learning techniques, $C$ is induced from a labeled dataset $D$ whose generic item is $(\bx, y_x)$, where $y_x$ equals $1$ if $\bx \in S$ and $0$ otherwise. In other words, $C$ is trained to classify keys in $S$ so that the higher is $C(\bx)$, the more likely $\bx \in S$. A binary prediction is ensured by thresholding using $\tau \in [0, 1]$, i.e.\ classifying $\bx \in U$ as a key if and only if $C(\bx) > \tau$.
\item Of course, nothing prevents us from having a set of false negatives $\{\bx \in S \, | \, C(\bx) \leq\tau\} \neq \emptyset$, thus a \emph{backup} (classical) Bloom Filter $F$ for this set is built. Summing up, $\bx \in U$ is predicted to be a key if $C(\bx) > \tau$, or $C(\bx) \leq \tau$ and $F$ does not reject $\bx$. In all other cases, $\bx$ is rejected.
\end{enumerate}

It is important to underline that the FPR of a classical BF is essentially independent of the distribution of data used to query it. This is no more true for a LBF~\cite{Mitz18}, in which such rate should be estimated from a query set $\overline S \subset U \backslash S$. To remark such difference, one commonly refers to the \emph{empirical FPR} of a learned filter, which is computed as $\epsilon = \epsilon_{\tau} + (1-\epsilon_{\tau})\epsilon_{F}$, where:
\begin{enumerate}
\item $\epsilon_{\tau} = |\{\bx \in \overline S \,|\,  C(\bx)> \tau\}|/|\overline S|$ is the analogous empirical FPR of the classifier $C$ on $\overline S$, and
\item $\epsilon_F$ is the false positive rate of the backup BF.
\end{enumerate}
Hence, having fixed a target value for $\epsilon$, the backup filter can be built setting $\epsilon_F = (\epsilon - \epsilon_{\tau})/(1 - \epsilon_{\tau})$ (under the obvious constraint $\epsilon_{\tau} < \epsilon$).
Within the learned setting, the three key factors of the filter are connected (and influenced) by the choice of $\tau$. Moreover, while data independence allows us to reliably estimate the FPR of a BF, this is no longer immediate for LBFs, as pointed out in~\cite{Mitz18}, along with an experimental methodology to assess it,  which is part of the   evaluation setting proposed in this paper.

Here below we outline the main features of the LBF variants which we have considered. With the exception of the one in \cite{Kraskap}, for which the software is neither public nor available from the authors, our selection is State of the Art.

\paragraph{Sandwiched LBFs~\cite{Mitz18}.} The Sandwiched variant of LBFs (SLBF for brevity) is based on the idea that space efficiency can be optimized by filtering out non-keys \emph{before} querying the classifier $C$, requiring as consequence a smaller backup filter $F$. More in detail, a BF $I$ for $S$ is initially built and used as a first processing step. All the elements of $S$ that are not rejected by $I$ are then used to build a LBF as described earlier. The SLBF immediately rejects an element $\bx \in U$ if $I$ rejects it, otherwise the answer for $\bx$ of the subsequent LBF is returned. The empirical FPR of the SLBF is  $\epsilon = \epsilon_I \big(\epsilon_{\tau} + (1-\epsilon_{\tau})\epsilon_{F} \big)$, where $\epsilon_{I}$ is the FPR of $I$. Here, fixed the desired $\epsilon$, the corresponding FPR to consctruct $I$ is $\epsilon_I = (\epsilon / \epsilon_{\tau}) (1-\mathrm{FN}/n)$, where $\mathrm{FN}$ is the number of false negatives of $C$.
Also in this case, the classifier accuracy affects the FPR, space and reject time, with the constraint $\epsilon ( 1-\mathrm{FN}/n ) \leq \epsilon_{\tau} \leq 1-\mathrm{FN} / n$.

\paragraph{Adaptive LBFs~\cite{Dai}.}
Adaptive LBFs (ADA-BF) represent an extension of LBF,  partioning the training instances $\bx$ into into $g$ groups, according to their classification score $C(\bx)$. Then, the same number of hash functions the backup filter of an LBF would use are partitioned across groups, and the membership for the instances belonging to a given group is tested only using the hash functions assigned to it.
As for the LBFs mentioned earlier, the expected FPR can only be estimated empirically. However, in this case the formula is rather complicated: the interested reader can refer to~\cite{Dai}.
The best performing variant of ADA-BF has been retained in this study.

\paragraph{Hyperparameters.}
The Learned Bloom Filters described above have some parameters to be tuned. Namely, the threshold $\tau$ for LBF and SLBF, and two parameters $g$ and $\bar c$ for ADA-BF, representing the number of groups in which the classifier score interval is divided into, and the proportion of non-keys scores falling in two consecutive groups. The details on the tuning of these hyperparameters are discussed in Section~\ref{sec:model_selection}.

\section{Experimental Methodology}\label{sec:methodology}
In this section we present the methodology which we adopt in order to design and analyse Learned Bloom Filters with regard to the inherent complexity of input data to be classified, subsumed as follows. The starting point is a dataset, either real-world or generated through a procedure suitable for synthesize data in function of some classification complexity metrics.

Overall, the pipeline adopted is as it follows: collect/generate data; induce a classifier from data and estimate its empirical FPR; construct a Learned Bloom Filter exploiting the learnt classifier, and in turn estimate its empirical FPR. The following sections review the considered classifier families and describe in depth the adopted data generation procedure.

\subsection{A Representative Set of Binary Classifiers}\label{sub:classifiers}

Starting from an initial list of classifiers---without presuming to be exhaustive---we performed a set of preliminary experiments, from which we received indications about the classifiers' families to be further analyzed, based on their trade-off between performance and space requirements\footnote{The experiments and data about this preliminary part are available upon request.}.   Namely, from the initial list we have removed the following classifiers: Logistic Regression~\cite{LRegr}, Naive Bayes~\cite{NaiveBC} and Recurrent Neural Networks~\cite{GRU}, due to their poor trade-off performance, confirming the results of a preliminary study~\cite{Fumagalli:2021}. The remaining ones are briefly described in the following paragraphs. Since our evaluation considers both balanced and unbalanced classification problems, we also detail how their inference is managed in an unbalanced context.
The hyperparameters of the corresponding learning algorithms are distinguished between \emph{regular} and \emph{key} hyperparameters, the latter affecting the space occupancy of the classifier. 
The model selection phase only involves non-key hyperparameters, while different configurations for key hyperparameters are analysed in dedicated experiments aiming at studying the interplay among FPR, space occupancy and reject time of Learned Bloom Filters.
Hereafter, when not differently specified, the space occupancy serves as a proxy for the complexity of a classifier.

In order to fix notation, we assume $U = \mathbb R^q$ as universe, and we refer to $D\subset U$ as a set of $d$ labeled instances, denoting by $\bx \in D$ a training instance and by $y_x \in \{ 0, 1 \}$  its label (with a slightly different notation for SVMs, as explained here below).

\paragraph{Linear SVM} Classification in Linear Support Vector Machines (SVMs) for a given instance $\bx$ is usually based on the optimal hyperplane ($\mathbf w, b$) and the sign of $f(\bx) = \mathbf w \cdot \bx + b$. To fall in the setting defined for a LBF, we need to transform the SVM prediction into a score in $[0,1]$. To this end, we use $f(\bx)$ as argument to a sigmoid function.
The optimal hyperplane ($\mathbf w, b$) is learned via maximization of the margin between the two classes, by solving
\begin{equation*}
\begin{array}{ll}
\min_{\mathbf w, b} & \frac{1}{2}\|\mathbf w\|^2 +\ c\sum_{\bx\in D}\xi_x~, \\
\text{such that} & y_xf(\bx) \geq 1-\xi_x  \quad \forall \bx \in D~, \\
& \xi_x \geq 0 \quad \forall \bx \in D~,
\end{array}
\end{equation*}
\noindent where misclassification is allowed by the slack variables $\xi_x$ and penalized using the hyperparameter $c>0$ (note that in this case $y_x \in \{-1, 1\}$). Nonlinear SVMs might have been used here, but the need of storing the kernel matrix, e.g., a Gaussian kernel, alongside the hyperplane parameters results in an unacceptable size for the learned filters. For the linear case we have chosen, we have only one non-key hyperparameter, namely $c$. When dealing with unbalanced labels, we consider the cost-senstive SVM version described in \cite{morik1999combining}.

\paragraph{Feed-Forward NNs} We also consider Feed-Forward neural networks~\cite{FFNN,haykin94} accepting instances as inputs and equipped with $l$ hidden layers, respectively having $h_1, \ldots, h_l$ hidden units
(NN-$h_1, \dots, h_l$ for short).
One output unit completes the network topology.
As usual, training involves the estimation of unit connections and biases from data. In this case, the $h_i$s are key hyperparameters, while the learning rate $lr$ acts as a regular hyperparameter to be tuned. It is worth noting that we fix the activation functions for all network units (although the former are tunable in principle), to limit the size of the already massive set of experiments. Precisely, as typically done, we use ReLU and Sigmoid activations for hidden and output units, respectively.
Where appropriate, label imbalance is dealt using a cost-sensitive model variant~\cite{BRUZZONE1997}.

\paragraph{Random Forests}
Finally, we consider also Random Forests~\cite{RF}, shortened as RF-$t$ to make explicit that this model is an ensemble of $t$ classification trees~\citep{CART}. Each such tree is trained on a different bootstrap subset of $D$ randomly extracted with replacement. Analogously, the splitting functions at the tree nodes are chosen from a random subset of the available attributes. The RF aggregates classifications uniformly across trees, computing for each instance the fraction of trees that output a positive classification. To address the case where labels are unbalanced, we adopt an imbalance-aware variant of RFs~\citep{VanHulse07,Khalilia11}
in which, during the growth of each tree, the bootstrap sample is not drawn uniformly over $D$, but by selecting an instance $\bx$ with probability
\[
\label{eq:weighting-scheme}
p_x = \begin{cases}
    \frac{1}{2|D_+|} & \text{if  $y_x=1$,} \\
	\frac{1}{2|D_-|} & \text{if $y_x=0$,}
\end{cases}
\]
where $D_+ = \{\bx \in D | y_x = 1\}$, and $D_- = D\setminus D_+$. In this way, the probabilities of extracting a positive or a negative example are both $1/2$, and the trees are trained on balanced subsets. The key hyperparameter of a RF is $t$, directly impacting on the classifier size. The non-key hyperparameter which we consider here is the minimum number $\delta$ of samples in a leaf, which allows to control the depth of the individual trees. It is worth observing that this hyperparameter can also slightly affect the tree size, but in our setting  the adopted grid of values only yields a negligible difference in size (see Section~\ref{sec:model_selection}).

\subsection{Measures of Classification Complexity and a Related Data Generation Procedure}\label{sec:measures}

In order to measure the complexity of a dataset, several measures are available, e.g. see~\cite{lorena2019} for a survey. We specifically focus on measures suitable for binary classification tasks, and hereafter we use the notation ``class $i$'', $i = 1, 2$, to refer to one of the two classes.
A preliminary analysis highlighted that some of the measures in~\cite{lorena2019}  were insensitive across a variety of synthetic data, as happened, e.g., with the \textit{F1}, \textit{T2}, or \textit{T3} measures, or needed an excessive amount of RAM (such as network- or neighborhood-based measures, like \textit{LSC} and \textit{N1}).
As a consequence, we selected the \textit{feature-based} measure
\textit{F1v} and the \textit{class-imbalance} measure \textit{C2}.
The former quantity, also called the Directional-vector Maximum Fisher's Discriminant Ratio, is defined as follows: denote, respectively, by $p_i$, $\mu_i$, and $\Sigma_i$ the proportion of examples belonging to class $i$ and the corresponding centroid and scatter matrix, so that $\boldsymbol{W} = p_1 \Sigma_1+ p_2 \Sigma_2$ and $\boldsymbol{B} = (\mu_1 - \mu_2)(\mu_1 - \mu_2)^\top$ are the between- and within-class scatter matrices. In this case, $\boldsymbol{d} = \boldsymbol{W}^{-1} (\mu_1 - \mu_2)$ corresponds to the direction onto which there is maximal separation of the two classes (being $\boldsymbol{W}^{-1}$ the pseudo-inverse of $\boldsymbol{W}$), and we can define the $F1v$ measure as
\begin{equation}
F1v = \left( 1 +
  \frac{\boldsymbol{d}^\top \boldsymbol{W} \boldsymbol{d}}
  {\boldsymbol{d}^\top \boldsymbol{B} \boldsymbol{d}}
  \right)^{-1}.
\end{equation}
The second measure accounts for label balance in the dataset: letting $n_i$ be the number of examples of class $i$, we have
$C2 = (n_1 - n_2)^2 / (n_1^2 + n_2^2)$.
Both measures vary in $[0,1]$: the higher the value, the more complex the data.

\begin{figure*}
\centering
\subfigure[]{\includegraphics[width=0.2\textwidth]{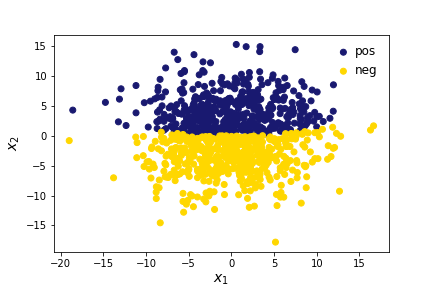}}
\subfigure[]{\includegraphics[width=0.2\textwidth]{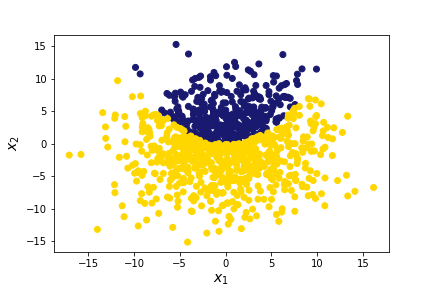}}
\subfigure[]{\includegraphics[width=0.2\textwidth]{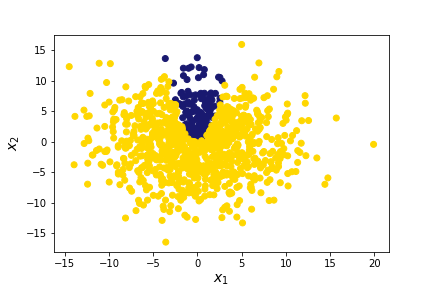}}\\
\subfigure[]{\includegraphics[width=0.2\textwidth]{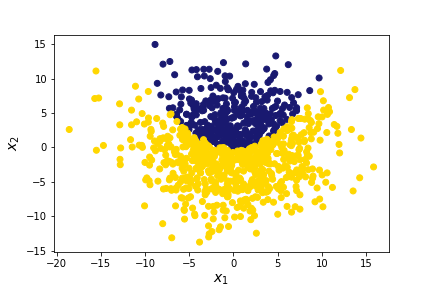}} \subfigure[]{\includegraphics[width=0.2\textwidth]{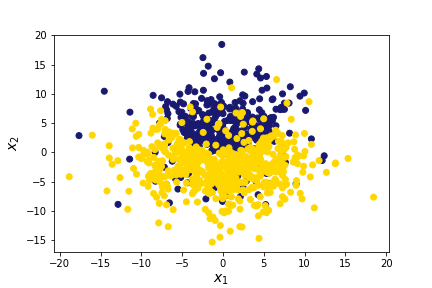}} \subfigure[]{\includegraphics[width=0.2\textwidth]{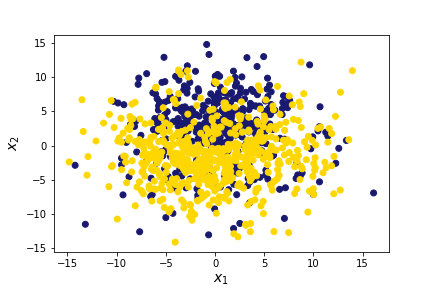}}\\
\subfigure[]{\includegraphics[width=0.2\textwidth]{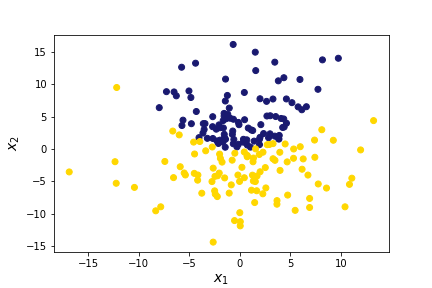}} \subfigure[]{\includegraphics[width=0.2\textwidth]{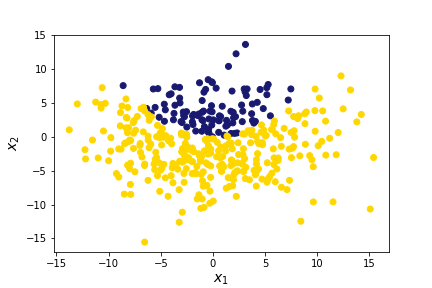}} \subfigure[]{\includegraphics[width=0.2\textwidth]{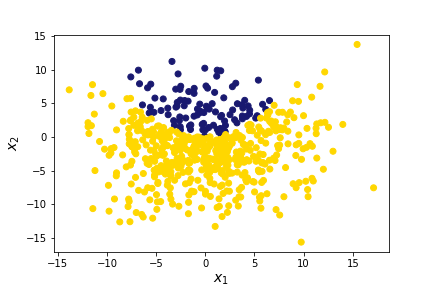}}


\caption{Graphical representation of synthetic data: first row, parameter configuration is $n_p=500$, $r=0$, $\rho =1$ and  $a=0.01$  (a), $a=0.1$ (b), and $a=1$ (c); second row $n_p=500$, $a=0.1$, $\rho =1$ and  $r=0$  (d), $r=0.1$ (e), and $r=0.25$ (f); third row, $n_p=100$, $a=0.1$, $r=0$, $\rho=1$  (g), $\rho=3$ (h), and $\rho=5$ (i). "pos" and "neg" entries in the legend stand for positive and negative class, respectively.}\label{fig:synt_data}
\end{figure*}

\paragraph{Data generation procedure.} We generate synthetic data considering three parameters, $a$, $r$ and $\rho$, which allow to tune the complexity of generated  data according to the aforementioned measures. Intuitively, $a$ controls the linearity of the separation boundary, $r$ the label noise, and $\rho$ the label imbalance.
More precisely, in order to generate a binary classification dataset with a given level of complexity, $n_1$ positive and $n_2=\lceil\rho n_1\rceil$ negative instances (with $N=n_1+n_2$), we proceed as follows.
Let $\{\bx_1, \dots \bx_N\} \subset \mathbb R^q$ be the set of samples, with each sample $\bx_i$ having $q$ features $x_{i1}, \ldots, x_{iq}$, and a binary label $y_i \in \{0, 1 \}$. The $N$ samples are drawn from a multivariate normal distribution $\mathcal{N}(0, \bSigma)$,  with $\bSigma = \gamma\bI_q$ (with $\gamma>0$ and $\bI_q$ denoting the $q \times q$ identity matrix). In our experiments we set $\gamma=5$ so as to have enough data spread, reminding that this value however does not affect the data complexity.
Without loss of generality,  we consider the case $q=2$. To determine the classes of positive and negative samples, the parabola $x_2 - ax_1^2 = 0$ is considered, with $a >0$: a point $\bx_i = (x_{i1},x_{i2})$ is positive ($y_i = 1$) if $x_{i2} - ax_{i1}^2 > 0$, negative otherwise ($y_i = 0$).
This choice allows us to control the linear separability of positive and negative classes by varying the parameter $a$: the closer $a$ to $0$, the more linear the separation boundary. As consequence, $a$ controls the problem complexity for a linear classifier.
An example of  generated data by varying $a$ is given in Figure~\ref{fig:synt_data} (a--c).
Further, to vary the data complexity even for non linear classifiers, labels are perturbed with different levels of noise: we flip the label of a fraction $r$ of positive samples, selected uniformly at random, with an equal number on randomly selected negatives. The effect of three different levels of noise is depicted in Figure~\ref{fig:synt_data} (d--f), where the higher the noise, the less sharp the separation boundary.
The third parameter $\rho$ is the ratio between the number of negative and positive samples in the dataset, thus it controls the C2 complexity measure. Higher values of $\rho$ make the negative boundary more clear (Figure~\ref{fig:synt_data} (g--i)), while making harder training an effective classifier~\cite{He09}.

\section{Experiments }\label{sec:reprod}
\subsection{Data}\label{sec:data}

\paragraph{Domain-Specific Data.} We use a URL dataset and a DNA dictionary. The first has been used by~\cite{Dai}, who also kindly provided us with the dataset, as part of their experimentation on Learned Bloom Filters. The second dataset comes  from experiments in  Bioinformatics regarding the storage and retrieval of $k$-mers (i.e., strings of length $k$  appearing in a given genome, whose spectrum is the dictionary of $k$-mers) \cite{Rahman21} and was directly generated by us. We point out that no sensible information is contained in these datasets.
With reference to Table~\ref{tab:real_complexity}, they represent two extreme cases of classification complexity: the URL dataset is \textit{easy}, as it is simple in terms of linear separability (F1v), albeit exhibiting a relevant C2 complexity due to the label imbalance; the DNA data is \textit{hard}, in that it has almost the maximum F1v possible value, meaning that positive and negative classes are indistinguishable by a linear classifier.

The URL dataset contains 485730 unique URLs, 80002 \textit{malicious} and the
remaining \textit{benign}. Seventeen lexical features are associated with each URL, which are used as the classification features. It is worth pointing out that all of the previous works on Learned Bloom Filters have used URL data. In this context, a Bloom filter can be used as a time- and space-efficient data structure to quickly reject benign URLs, never erroneously trusting a malicious URL although occasionally misclassifying a benign one. We adhere to this standard here.

The DNA dataset refers to the human chromosome 14, containing $n = 49906253$ 14-mers~\cite{Rahman21} constituting the set of our keys. As non-keys, we uniformly generate other $n$ 14-mers from the $4^{14}$ possible strings on the alphabet $\{A, T, C, G\}$.
Each 14-mer is associated with a 14-dimensional feature vector, whose components are the integers $0,1, 2, 3$, each associated with one of the four DNA nucleobases A, T, C, G, respectively (for instance a 14-mer TAATTACGAATGGT is coded as $(1,0,0,1,1,0, 2, 3, 0, 0, 1, 3, 3, 1)$). A fundamental problem in Bionformatics, both technological \cite{Solomo} and in terms of evolutionary studies \cite{chor2009genomic},  is to quickly establish whether a given $k$-mer belongs to the spectrum of a genome. In this case, the Bloom Filter stores the dictionary.  It is worth mentioning that the use of Bloom Filters in Bioinformatics is one of their latest fields of application, with expected high impact \cite{Elworth2020}. Such a domain has not been considered for the evaluation of Learned Bloom Filters, as we do here.

\paragraph{Synthetic Data.}
We generate two categories  of synthetic data, each attempting to reproduce the complexity of one of the domain-specific data.
The first category has nearly the same C2 complexity of the URL dataset, i.e., it is \textit{unbalanced}, with $n_1=10^5$ and  $\rho = 5$. The second one has the same C2 complexity of the DNA dataset, i.e., it is \textit{balanced}, with  $n_1=10^5$ and $\rho = 1$.
The choice of $n_1$ allows to have a number of keys similar to that in the URL data, and at the same time to reduce the computational burden of the massive set of experiments planned. Indeed, both balanced and unbalanced categories contain nine datasets, exhibiting increasing levels of $F1v$ complexity. Specifically, all possible combinations of parameters $a \in \{ 0.01, 0.1, 1 \}$ and $r \in \{ 0, 0.1, 0.25 \}$ are used. The corresponding complexity estimation is shown in Table~\ref{tab:compl}. Consistently, F1v complexity increases with $a$ and $r$ values, in both balanced and unbalanced settings. Noteworthy, the label imbalance slightly affects also the measure F1v: in absence of label noise ($r=0$),
F1v augments, likely due to the fact that F1v is an \textit{imbalance-unaware} measure; on the contrary, in presence of noise F1v complexity is barely reduced w.r.t. the balanced case.
Although not immediate, the sense of this behavior might reside in what we also observe in Figure~\ref{fig:synt_data}(g--i).  That is, the boundary of negative class tends to be more crisp when $\rho$ increases, thus mitigating the opposite effect the noise has on the boundary (Figure~\ref{fig:synt_data} (d--f)).

\begin{table}
    \caption{Complexity of the real data.}
    \label{tab:real_complexity}
    \centering
    \begin{tabular}{lcc}
        \toprule
        \textbf{Data} & \textbf{F1v} &\textbf{C2} \\
        \midrule
        URL &  0.08172 & 0.62040 \\
        DNA & 0.99972 & 0 \\
        \bottomrule
    \end{tabular}
\end{table}
\begin{table}
\caption{Complexity of the synthetic data.}\label{tab:compl}
\centering
\begin{tabular}{llcc|cc}
  \hline
 & & \multicolumn{2}{c|}{{Balanced}}&\multicolumn{2}{|c}{{Unbalanced}}\\[1pt]
$a$ & $r$ & \textbf{F1v} & \textbf{C2} & \textbf{F1v} & \textbf{C2} \\
  \hline
$0.01$ & $0$    & 0.127 & 0.0& 0.129 & 0.615 \\
$0.1$  & $0$    & 0.181 & 0.0 & 0.202 & 0.615 \\
$1$    & $0$    & 0.306 & 0.0& 0.360 & 0.615 \\
$0.01$ & $0.1$  & 0.268 & 0.0& 0.187 & 0.615 \\
$0.1$  & $0.1$  & 0.327 & 0.0& 0.269 & 0.615 \\
$1$    & $0.1$  & 0.459 & 0.0& 0.433 & 0.615 \\
$0.01$ & $0.25$ & 0.571 & 0.0& 0.308 & 0.615 \\
$0.1$  & $0.25$ & 0.619 & 0.0& 0.399 & 0.615 \\
$1$    & $0.25$ & 0.718 & 0.0& 0.563 & 0.615 \\
   \hline
\end{tabular}
\end{table}

\subsection{Hardware and Software}\label{sec:HW}
We use two Ubuntu machines: an Intel Core i7-10510U CPU at 1.80GHz$\times$8 with 16GB RAM, and an Intel Xeon Bronze 3106 CPU at 1.70GHz$\times$ 16 with 192GB RAM. This latter is used for experiments that require a large amount of RAM, i.e., on the DNA dataset.
The supporting software~\cite{supp-software} is written in Python 3.8, leveraging the ADA-BF public implementation provided in~\cite{adabf}, which we extend as follows:
\begin{inparaenum}
    \item the construction of Learned Bloom Filters can be done in terms of the classifiers listed in Section~\ref{sub:classifiers} and of the datasets illustrated in Section~\ref{sec:data};
    \item SLBF is added to the already included BF models;
    \item the choice of the classifier threshold $\tau$ is performed considering any number of evenly spaced percentiles of the obtained classification scores, instead than checking fixed values;
    \item ranges for the hyperparameters of the learned versions of BFs can be specified by the user;
    \item a main script allows to perform all experiments, rather than invoking several scripts, each dedicated to a LBF variant.
\end{inparaenum}

The provided implementation is built on top of standard libraries, listed in a dedicated environment file in order to foster replicability. In particular,
the space required by a given classifier is computed, as typically done in these cases, using the \emph{Pickle} module and accounting for both structure information and parameters~\cite{Pickle}, in order to obtain a fair comparison among all tested configurations. Moreover, the software is opened to extensions concerning the inclusion of new datasets and/or new LBF models, thus it can be used as a starting point for further independent researches.


\subsection{Model Selection}\label{sec:model_selection}
\paragraph{Classifiers.} The classifier generalization ability is first assessed independently of the filter employing it, via a $3$-fold cross validation (CV) procedure (outer). Classifier performance is measured in terms of \begin{inparaenum}[(i)]
\item the area under the ROC curve (AUC), and of
\item the area uder the precision-recall curve (AUPRC),
\end{inparaenum}
averaged across folds.
We tune non-key hyperparameters of each model via a nested $3$-fold  (CV), where in each round of the outer CV we select the non-key hyperparameters through a grid search using the inner CV on the current training set, and the best configuration is retained.  We use the following grids: $c \in \{10^{-1}, 1, 10, 10^2, 10^3\}$ (SVM); $\delta \in \{1, 3, 5\}$ (RF); $lr \in \{10^{-4}, 10^{-3}\}$ (NN). The different size of the grid across classifiers is due to the different training time of the classifier (SVM is the fastest one).
The configuration of a classifier is strictly dependent on the space budget assigned to the LBF leveraging that classifier (see Table~\ref{tab:budget} discussed in next section); consequently, the key hyperparameters for a given classifier, i.e., hyperparameters influencing the space occupancy,  are set based on the following strategy: to detect the subspace of ``valid'' configurations, we carried out several preliminary experiments (not shown here) for each classifier, subject to the available space budget, and also evaluating their performance/space trade-off. The results described here below leverage this preliminary filter. Recalling that no key hyperparameters exist for SVMs, we consider RFs related to two values of $t$, leading to a simpler and a more complex model. The simpler choice is $t=10$, as a  reference already used in the literature~\cite{Dai}, whereas we consider $t=20$ and $t=100$ as more complex variants, used respectively for the synthetic/URL and for the DNA datasets (cfr.\ Section~\ref{sec:data}). This distinction is due to different key set size, in turn influencing the available budget, and the choice $t=100$ needs more space than the  budget available for URL and synthetic data.
For a fair comparison, the key hyperparameters for NNs are selected so as to yield three models nearly having the same occupancy of the SVM and of the two RFs models. The above-mentioned preliminary experiments have suggested, where enough space budget was available, that a two-layered topology is to be preferred to the one-layered one. Precisely, we consider the following models: NN-$25$, NN-$150,50$ and NN-$200,75$ (synthetic dataset); NN-$7$, NN-$150,35$ and NN-$175,70$ (URL dataset); NN-$7$, NN-$125,50$, NN-$500,150$ (DNA dataset). The final classifier configuration for all experiments and their space requirements are detailed in Table~\ref{tab:cl_space}.
For the final discussion, in Table~\ref{tab:cl_time} we also include the average prediction time of the tested classifiers.
\begin{table*}
\caption{Space occupancy in Kbits of selected classifiers on the considered datasets.}\label{tab:cl_space}
\centering
\begin{tabular}{llllll}
  \hline
\multicolumn{6}{c}{{\tt Synthetic Data}}\\[2pt]
\textbf{SVM} & \textbf{RF-10} & \textbf{RF-20} & \textbf{NN-25} & \textbf{NN-150,50}& \textbf{NN-200,75}\\
5 & 259.3 & 508.6 & 5.1 & 260.2 & 506.6\\
\multicolumn{6}{c}{{\tt URL Data}}\\[2pt]
\textbf{SVM} & \textbf{RF-10} & \textbf{RF-20} & \textbf{NN-7} & \textbf{NN-150,35}& \textbf{NN-175,70}\\
5.9 & 259.3 & 508.7 & 6.2 & 259.2 & 499.9\\
\multicolumn{6}{c}{{\tt DNA  Data}}\\[2pt]
\textbf{SVM} & \textbf{RF-10} & \textbf{RF-100} & \textbf{NN-7} & \textbf{NN-125,50}& \textbf{NN-500,150}\\
5.8 & 259.5 & 2504 & 5.6 & 265.8 & 2652.3\\
   \hline
\end{tabular}
\end{table*}
\begin{table*}
\caption{Average classifier inference time (across samples) in seconds.}\label{tab:cl_time}
\centering
\begin{tabular}{llllll}
  \hline
\multicolumn{6}{c}{{\tt Synthetic Data}}\\[2pt]
\textbf{SVM} & \textbf{RF-10} & \textbf{RF-20} & \textbf{NN-25} & \textbf{NN-150,50}& \textbf{NN-200,75}\\
$1.278\cdot 10^{-8}$ & $4.425\cdot 10^{-7}$ & $8.968\cdot 10^{-7}$ & $8.494\cdot 10^{-6}$ & $9.257\cdot 10^{-6}$ & $1.008\cdot 10^{-5}$\\
\multicolumn{6}{c}{{\tt URL Data}}\\[2pt]
\textbf{SVM} & \textbf{RF-10} & \textbf{RF-20} & \textbf{NN-7} & \textbf{NN-150,35}& \textbf{NN-175,70}\\
$3.730\cdot 10^{-8}$ & $5.815\cdot 10^{-7}$ & $9.930\cdot 10^{-7}$ & $6.825\cdot 10^{-6}$ & $7.018\cdot 10^{-6}$ & $7.198\cdot 10^{-6}$\\
\multicolumn{6}{c}{{\tt DNA  Data}}\\[2pt]
\textbf{SVM} & \textbf{RF-10} & \textbf{RF-100} & \textbf{NN-7} & \textbf{NN-125,50}& \textbf{NN-500,150}\\
$2.87\cdot 10^{-8}$ & $5.572\cdot 10^{-7}$ & $5.364\cdot 10^{-6}$ & $6.572\cdot 10^{-6}$ & $8.138\cdot 10^{-6}$ & $1.044\cdot 10^{-5}$\\
   \hline
\end{tabular}
\end{table*}
\begin{table*}
\caption{Space budget in bits adopted on the various datasets. $\epsilon$ is the false positive rate, $n$ is the number of keys in the dataset.}\label{tab:budget}
\centering
\begin{tabular}{llll}
  \hline
\textbf{Data}& $\boldsymbol{\epsilon}$ & \textbf{Budget (Kbits)} & $\boldsymbol{n}$\\
Synthetic & $0.05, 0.01$ & 622, 956 & $10^5$\\
URL & $0.01, 0.005, 0.001, 0.0005, 0.0001$ & $765, 880, 1148, 1263, 1530$ & $8\cdot 10^4$\\
DNA & $0.01, 0.005, 0.001$, $0.0005$, $0.0001$
& $477460, 549325, 716191$, $788056$, $954921$
& $4.99 \cdot 10^7$\\
   \hline
\end{tabular}
\end{table*}

\paragraph{Learned Bloom Filters.}
The Bloom Filter variants under study are evaluated under the setting proposed by~\cite{Dai}, that is:
\begin{inparaenum}
    \item train the classifiers on all keys and $30\%$ of non-keys, and query the filter using remaining $70\%$ of non-keys to compute the empirical FPR;
    \item fix an overall memory budget of $m$ bits for each filter, and compare them in terms of their empirical FPR $\epsilon$.
\end{inparaenum}
In addition, in this work we also measure the \textit{average reject time} of filters, since it is can unveil interesting trends about the synergy of the filter variants and the classifier they employ. Each filter variant is trained leveraging in turn each of the considered classifiers.
The budget $m$ is related to the desired (expected) $\epsilon$ of a classical Bloom Filter, according to~(\ref{eq:bloom}). Being the space budget directly influenced by the key set size $n$, we adopt a setting tailored on each dataset.
Concerning synthetic data, as we generate numerous datasets, for each of them we only test two different choices for the space budget $m$. Namely, those yielding $\epsilon \in \{ 0.05, 0.01 \}$ for the classical Bloom Filter using a bit vector of $m$ bits according to (\ref{eq:bloom}).
On real datasets we test five budgets corresponding to $\epsilon \in \{ 0.01, 0.005, 0.001, 0.0005, 0.0001 \}$. The difference between this setting and that of synthetic data is due to the following considerations. First, the dimensionality of synthetic data is $2$, whereas that of real data it is $17$ and $14$, respectively, for URL and DNA. This makes the classifiers using real data larger than their counterparts on synthetic data. For this reason, on real data we omit the case $\epsilon = 0.05$, which yielded a too small budget. Indeed, some classifiers alone exceed the budget in this case (cfr.\ Table~\ref{tab:cl_space} for details about the space occupancy of classifiers). Moreover, having only two datasets, we can test more choices of $\epsilon$, and accordingly better evaluate the behavior of learned Bloom Filters when a smaller (expected) false positive rate is required.
Table~\ref{tab:budget} contains the resulting budget configurations for all the considered datasets.
To build the learned Bloom Filters variants, the hyperparameters have been selected via grid search on the training data, optimizing with respect to the FPR, according to the following setting: \begin{inparaenum}[(a)]
    \item $15$ different values for threshold $\tau$, and
    \item the ranges $[3, 15]$, and $[1, 5]$ for hyperparameters $g$ and $\bar c$, respectively (cfr.\ Section~\ref{sec:LBF}).
\end{inparaenum}
Importantly, the latter choice includes and extends the configurations adopted in~\cite{adabf} (namely, $[8,12]$ for $g$ and $[1.6, 2.5]$ for $\bar c$).
\begin{figure*}
\centering
\subfigure{\includegraphics[width=0.35\textwidth]{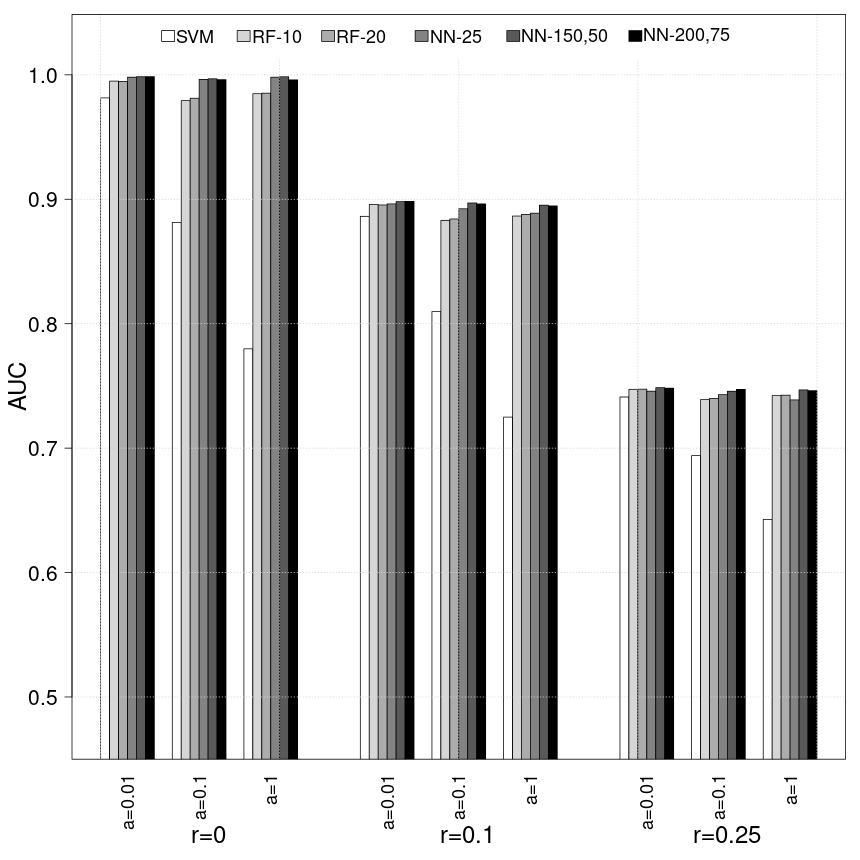}}
\subfigure{\includegraphics[width=0.35\textwidth]{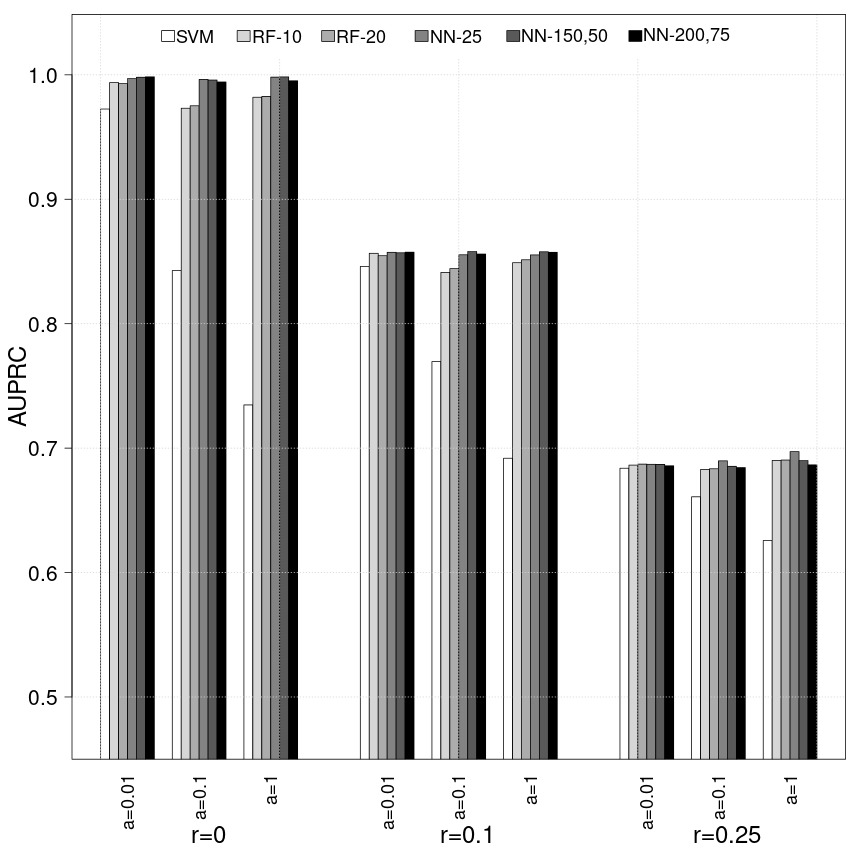}}\\
\subfigure{\includegraphics[width=0.35\textwidth]{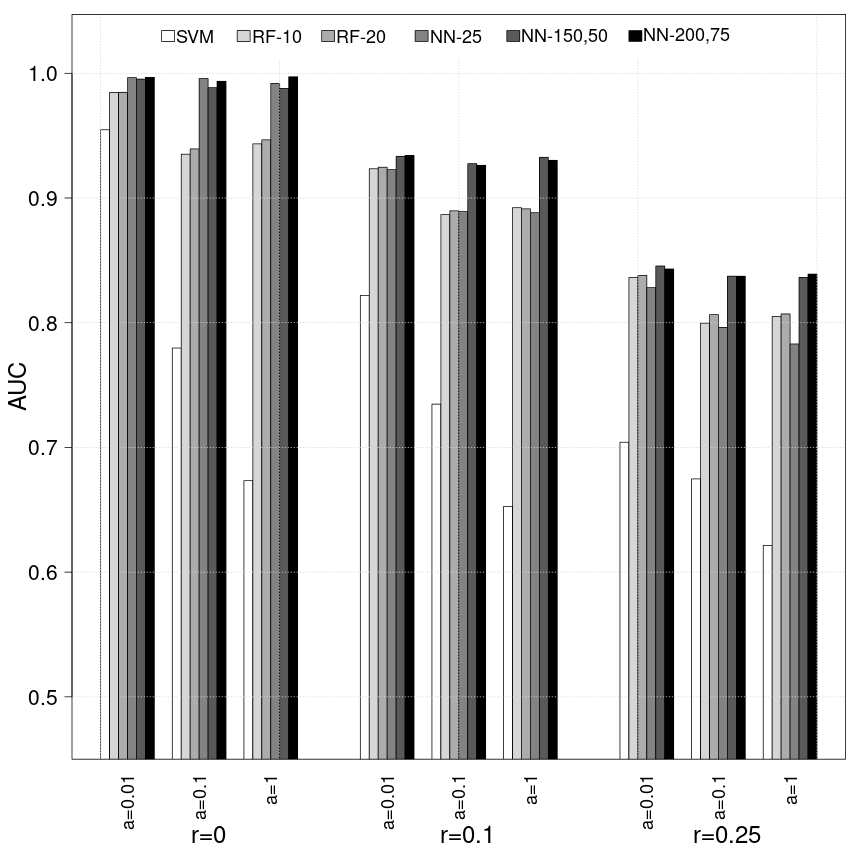}}
\subfigure{\includegraphics[width=0.35\textwidth]{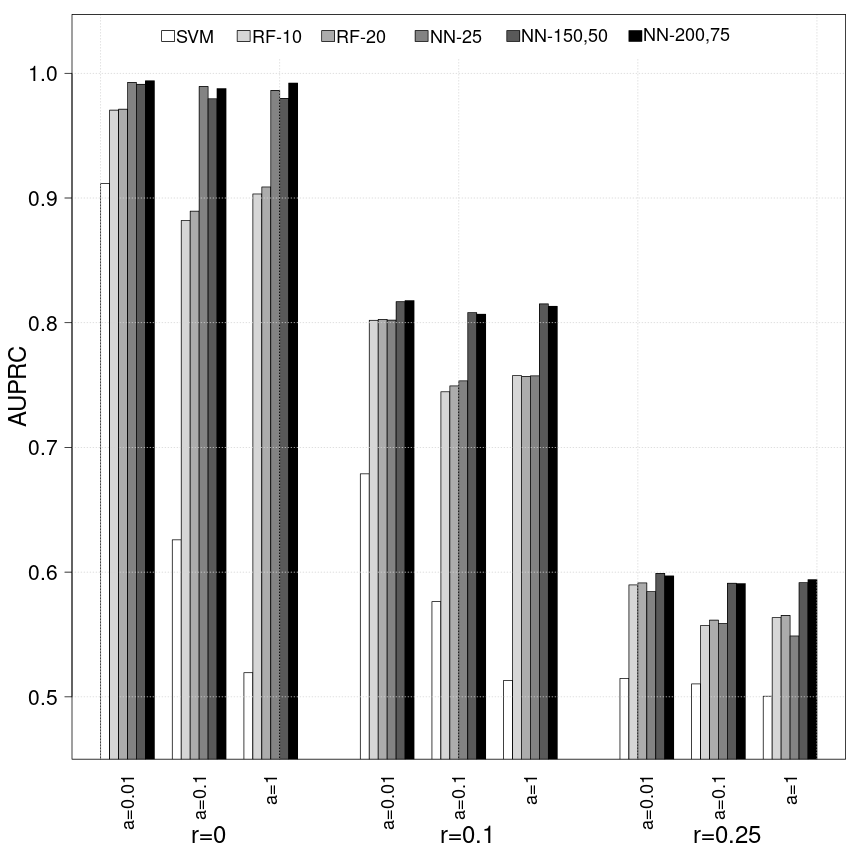}}

\caption{Performance averaged across folds of compared classifiers on synthetic data. First row for \textit{balanced} data, second row for \textit{unbalanced} data. Bars are grouped by dataset, in turn denoted by a couple ($a$, $r$).}\label{fig:results_synth}
\end{figure*}

\begin{figure*}
\centering
\subfigure[]{\includegraphics[width=0.27\textwidth]{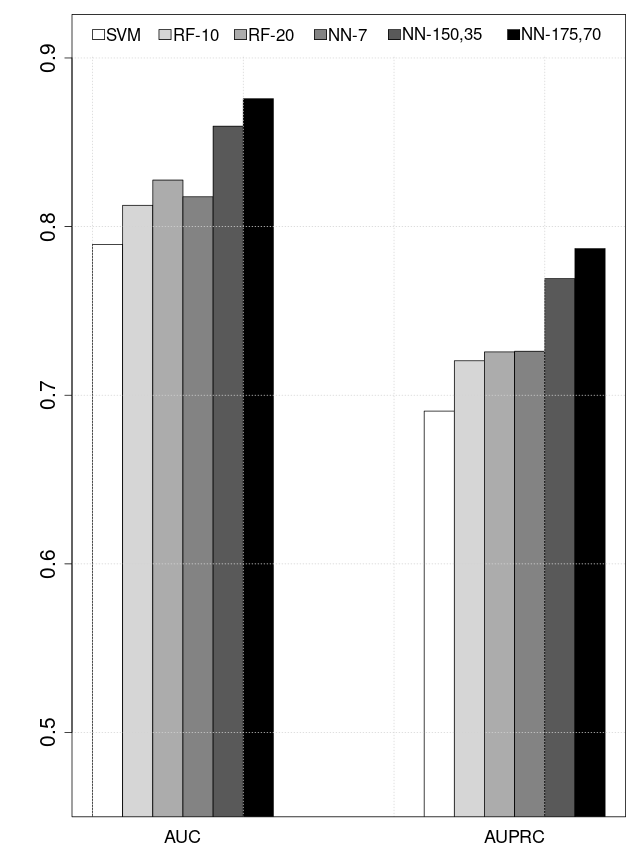}}
\subfigure[]{\includegraphics[width=0.27\textwidth]{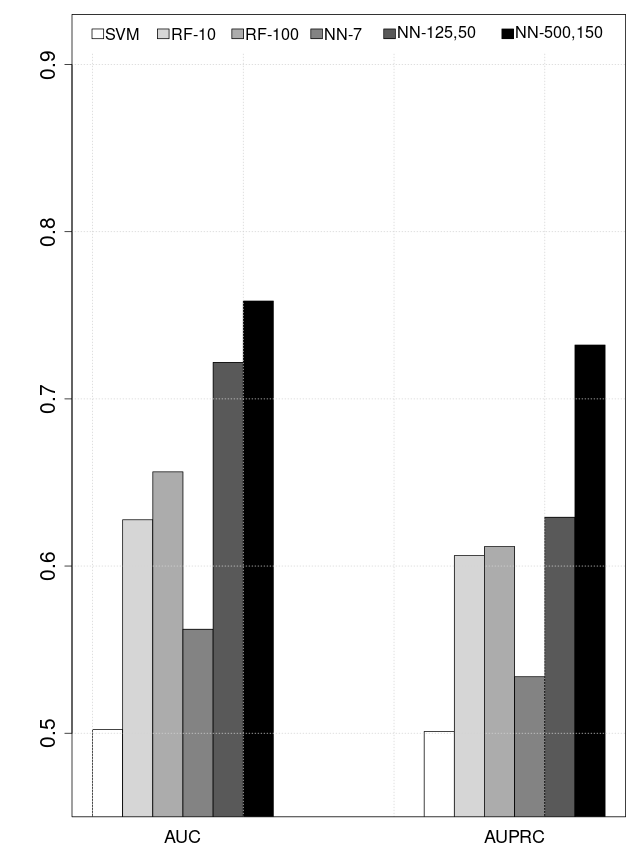}}
\caption{Performance averaged across folds of compared classifiers on real data: (a) URL; (b) DNA.}\label{fig:results_real}
\end{figure*}
\section{Results and Discussion}\label{sec:complex}
In this section  we present the results obtained from the classifier screening on both synthetic
and real data, the experimental evaluation of all variants of LBFs based on those
classifiers, and the relative discussion.
\subsection{Classifiers}\label{sub:class_discuss}
As evident from Section~\ref{sec:LBF}, the classifier can be interpreted as an oracle for a learned BF, where the better the oracle, the better the related filter, i.e., its FPR once fixed the space budget. Accordingly, it is of interest to evaluate the performance of classifiers.
All classifiers described in Section~\ref{sub:classifiers} have been tested on the datasets  described  in Section~\ref{sec:data}, with the configuration described in Section \ref{sec:model_selection}.
 Figure~\ref{fig:results_synth}  depicts the performance of classifiers on \textit{balanced} and \textit{unbalanced} synthetic data, whereas results obtained  on real data are shown in Figure~\ref{fig:results_real}.
However, it is central here to emphasize that the interpretation of such results is somewhat different than what one would do in a standard machine learning setting. Indeed, we have a space budget for the entire filter, and the classifier must discriminate well keys and non-keys, while being substantially succinct with regard to the space budget of the data structure. Such a scenario implicitly imposes a  performance/space trade-off: hypothetically, we might have a perfect classifier using less space than the budget, and on the other extreme, a poor classifier exceeding the space budget.
\subsubsection{Overall Results Analysis}\label{subsub:ORA}
First, the behaviour of classifiers in  terms of AUC and AUPRC is coherent with what expected according to our methodology to generate synthetic data. Indeed, the SVM performance decays when parameter $a$ increases, being in line with the fact that it means increasing the non-linearity of the class separation boundary. Analogously, all classifiers worsen as noise $r$ increases, which is clearly  what to expect in this case.
Moreover and most importantly, two main cases arise with respect to classification complexity: roughly F1v $\leq 0.35$ and F1v $> 0.35$. Being this threshold experimentally derived, the division between the two cases is not crisp.
We refer to the first case as datasets `easier and easier' to classify, for brevity `easy', and to the second as datasets `harder and harder' to classify, for brevity `hard'.
\paragraph{Easy datasets.}
All classifiers perform very well on synthetic datasets with the stated complexity (except for SVMs when $a>0.01$).
Clearly, with such excellent oracles, the remaining part of a learned Bloom Filter (e.g., with reference to the description of LBF, the backup filter) is intuitively expected to be very succinct.

\paragraph{Hard datasets.}
In this case, both AUC and AUPRC sensibly drop, being in some cases (SVM) not so far from the behaviour of a random classifier.
While in the previous case the performance of classifiers clearly yields the choice of the most succinct and faster model, here there is a trade-off to consider.
Indeed, within the given space budget,  at one end of the spectrum, we have the choice of a small-space and inaccurate classifier, at the other end of the spectrum
we have larger and more accurate ones.
As an example, for the LBF in the first case  a large backup filter is required, whereas in the second one the classifier would use most of the space budget.
\subsubsection{Preliminary observations on the classifiers to be retained.}
Here we address the question of how to choose a classifier to build the filter upon, based only on the knowledge of space  budget and data classification complexity/classifier performance.
On synthetic and URL data (Figures~~\ref{fig:results_synth} and~\ref{fig:results_real} (a)), more complex classifiers perform just slightly better than the simpler ones, likely due to the low data complexity in these cases. At the same time, they require a sensibly higher fraction of the space budget (Table~\ref{tab:budget}), and it is thereby natural to retain in this cases only the smallest/simplest variants, namely: RF-$10$ and NN-$25$ (synthetic) and NN-$7$ (URL), in addition to SVM. Conversely, in DNA experiments more complex classifiers substantially outperform the simpler counterparts, coherently with the fact that this classification problem is much harder  (Tables~\ref{tab:real_complexity} and~\ref{tab:compl}). Since the available space budget is higher in this case, all classifiers have been retained in the subsequent filter evaluation.
\subsection{Learned Bloom Filters Evaluation}\label{sec:experiments}
The aim of this section is:
\begin{inparaenum}
\item to explore how the various learned filters behave with respect to the data classification complexity, an aspect so far ignored in the literature (see Introduction);
\item to include the reject time in the overall filter assessment;
\item to gain further insights about the interplay between  the different classifiers and the learned Bloom Filter variants.
\end{inparaenum}

\begin{figure*}
\centering
\subfigure{\includegraphics[width=0.25\textwidth]{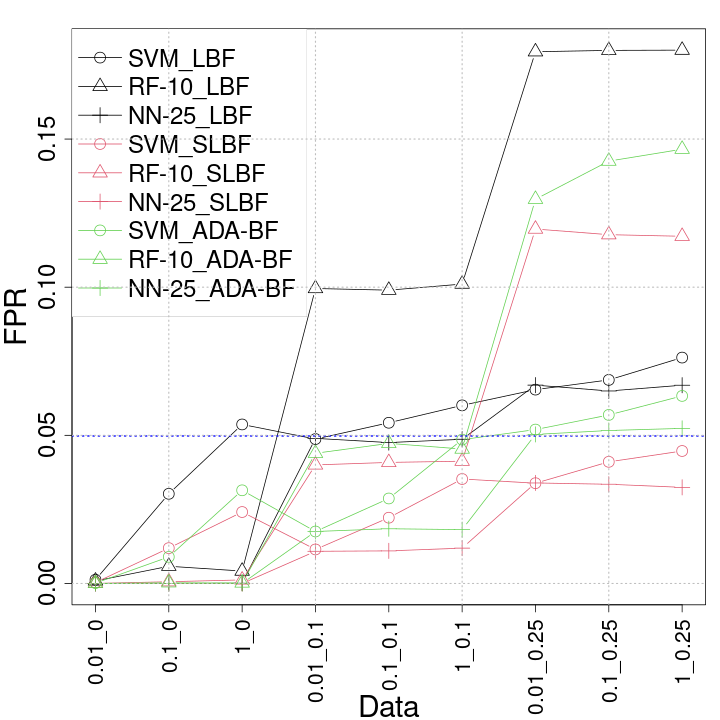}}
\subfigure{\includegraphics[width=0.25\textwidth]{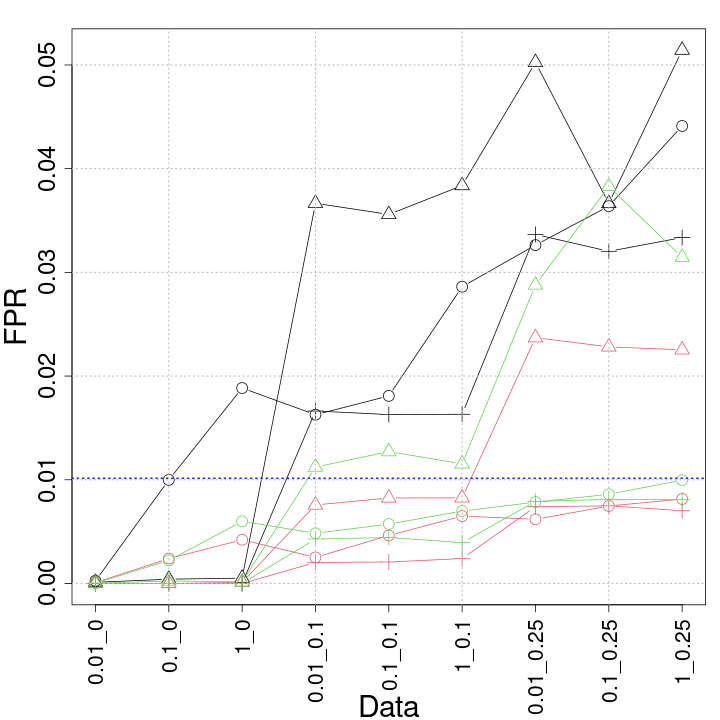}}
\caption{False positive rates of learned filters attained on balanced synthetic datasets. On the horizontal axis, labels $X\_Y$ denote the dataset obtained when using $a = X$ and $r=Y$. The blue dotted line corresponds to the empirical false positive rate of the classical BF in that setting. Two space budgets $m$ are tested, ensuring that $\epsilon = 0.05$ (left) and $\epsilon = 0.01$ (right) for the classical BF .}\label{fig:filter_synth_bal}
\end{figure*}

\begin{figure*}
\centering
\subfigure{\includegraphics[width=0.25\textwidth]{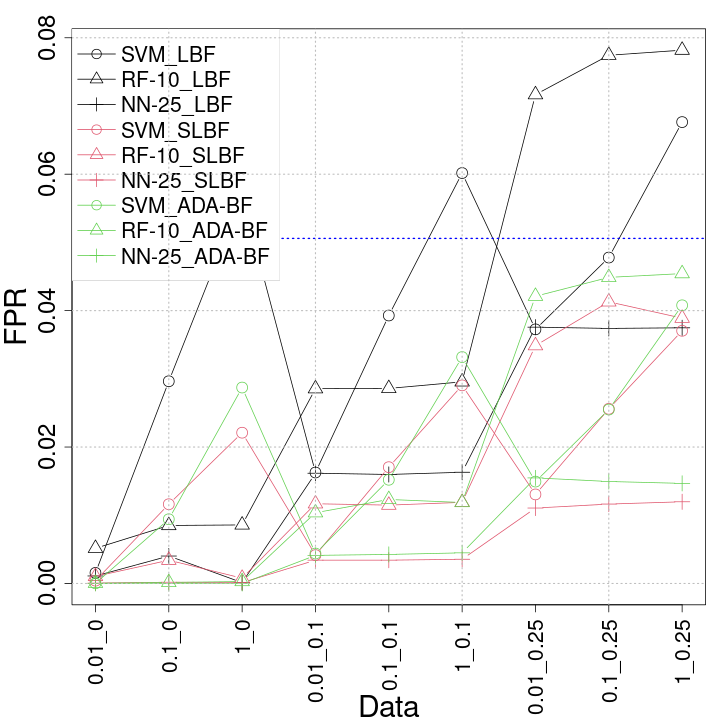}}
\subfigure{\includegraphics[width=0.25\textwidth]{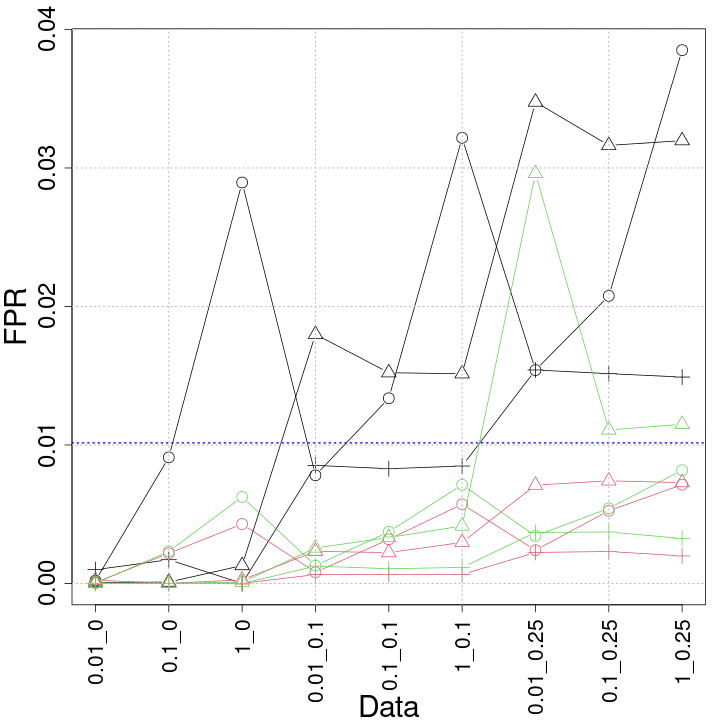}}
\caption{False positive rates of learned filters attained on unbalanced synthetic datasets. On the horizontal axis, the labels $X\_Y$ denote the dataset obtained when using $a = X$ and $r=Y$. The blue dotted line corresponds to the measured false positive rate of the classical Bloom filter in that setting. Two space budgets are tested: that ensuring $\epsilon = 0.05$ for a classical Bloom Filter (left), and that ensuring $\epsilon = 0.01$ (right).}\label{fig:filter_synth_unbal}
\end{figure*}


\subsubsection{Learned Filters Performance and Their Relationship with Data Classification Complexity.}\label{subsub:filtperf}
\begin{figure*}
\centering
\subfigure{\includegraphics[width=0.25\textwidth]{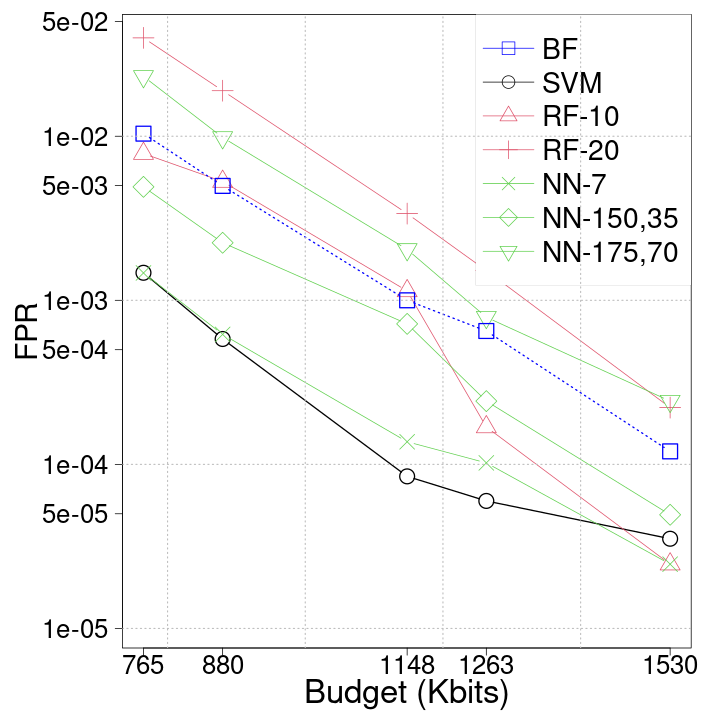}}
\subfigure{\includegraphics[width=0.25\textwidth]{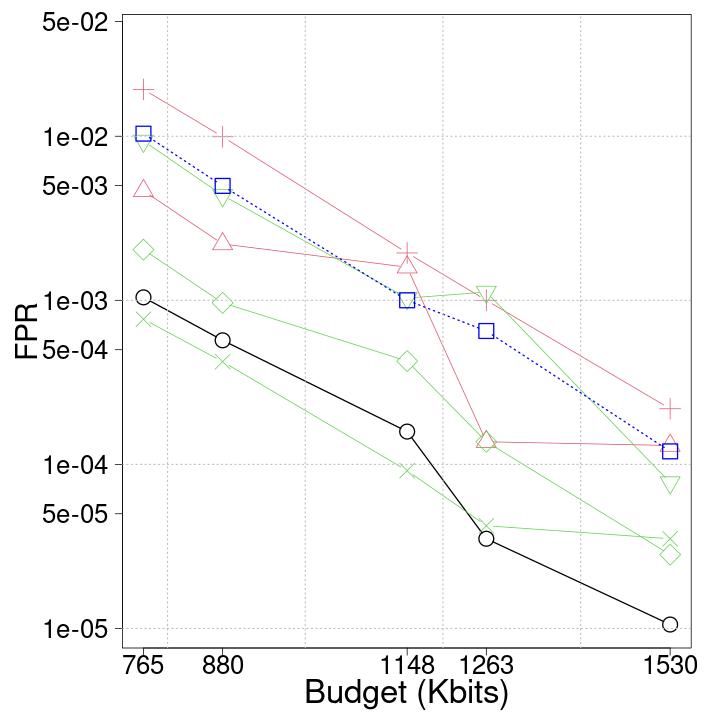}}
\subfigure{\includegraphics[width=0.25\textwidth]{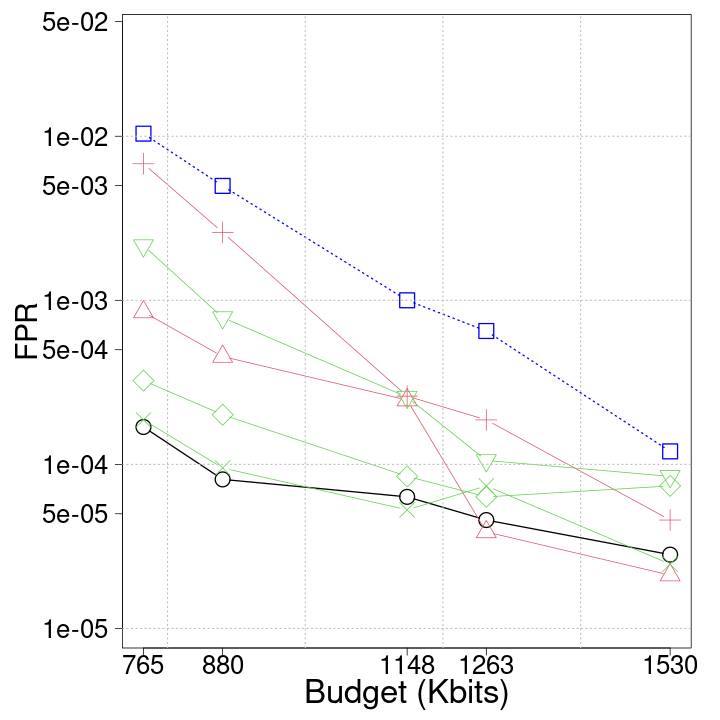}}
\caption{Empirical false positive rate of LBF (left), SLFB (central), and ADA-BF (right) filters on URL data. On the horizontal axis the different budgets configurations. Dotted blue line represents the baseline classical Bloom Filter. }\label{fig:filters_URL}
\end{figure*}

\paragraph{Easy datasets.}
Figures~\ref{fig:filter_synth_bal} and~\ref{fig:filter_synth_unbal} report the FPR results of learned Bloom Filters on balanced and unbalanced synthetic data, respectively, whereas Figures~\ref{fig:filters_URL} and~\ref{fig:filters_DNA} depict the results on URL and DNA data.
In all figures, also the baseline Bloom Filter is present.
According to the definition provided in Section~\ref{subsub:ORA} (F1v around $0.35$ or smaller), easy data can be associated to the three/four leftmost configurations on the $x$-axis in Figures~\ref{fig:filter_synth_bal} and~\ref{fig:filter_synth_unbal} of synthetic and URL data.
In these cases, we observe results coherent with those obtained in the literature, where ADA-BF slightly outperforms the other competitors~\cite{Dai}, and RF-$10$ inducing however lower FPR values with regard to the classical BF. Notwithstanding, it clearly emerges that such a classifier is not the best choice, underlining all the doubts about a selection not motivated in the original studies. For instance, on URL data there are at least two classifiers yielding a lower FPR in most  cases and for all filter variants (SVM and NN-$7$). In additon, SVMs are much faster (Table~\ref{tab:cl_time}). NN-$7$ (or NN-$25$ for synthetic data) remains the best choice even when the separation boundary becomes less linear ($a>0.01$), and filters induced by SVMs become less effective or even worse than the baseline BF.
\paragraph{Hard datasets.}
Our experiments show a novel scenario with the increase of data complexity, i.e., when moving towards right on the horizontal axis in Figures~\ref{fig:filter_synth_bal} and~\ref{fig:filter_synth_unbal}, or when considering DNA data. We observe that the performance of the filters drops more and more, in line with the performance decay of the corresponding classifiers (Section~\ref{sub:class_discuss}), and unexpectedly the drop is faster in ADA-BF (and LBF) w.r.t.\ SLBF. This happens for instance on all synthetic data having $r>0$ (noise injection). We say unexpectedly since we have an inversion of the trend also reported in the literature, where usually ADA-BF outperforms SLBF (which in turn improves LBF). Indeed SLBFs here exhibit behaviours more robust to noise,
which are likely due to a reduced dependency on the classifier for SLBF, yielded by the usage of the initial Bloom Filter. Such a filter indeed allows the classifier to be queried only on a subset of data. Noteworthy is the behavior of filters when using RFs in this setting: their FPR strongly increases, and potential explanations are the excessive score discretization (having 10 trees we have only $11$ distinct scores for all queries), and the space occupancy is larger (limiting the space to be assigned to initial/backup filters). These results 
find a particularly relevant confirmation on the very hard, real-world, large, and novel DNA dataset (Figure~\ref{fig:filters_DNA}). Here, surprisingly, the LBF cannot attain any improvement with regard to the baseline BF, differently from SLBF and ADA-BF.
A potential cause can reside in the worse performance achieved by classifiers on this hard dataset, compared to those obtained on synthetic and URL data, and in a too marked dependency of LBF on the classifier performance, mitigated instead in the other two filter variants by the usage of the initial BF (SLBF) and by the fine-grained classifier score partition (ADA-BF).
SLBF outperforms both LBF and baseline of one order of magnitude in FPR with the same space amount, and ADA-BF when using weaker classifiers and when a higher budget is available. This is likely due to overfitting of ADA-BF in partitioning the classifier codomain when the classifier performance is not excellent (or similarly when the data complexity is high), as it happens for DNA data. Differently from hard synthetic data, where the key set was smaller (and consequently also space budget was smaller), here the classifiers leading to the best FPR values are the most complex, in particular NN-500,150 and NN-125,50 (which are also the top performing ones, and might be further compressed~\cite{neucom}). In other words, it means that on hard datasets, simple classifiers are useless or even deleterious (SVM-induced filters never improve the baseline, and in some cases they are even worse).

\subsubsection{Reject time.}\label{subsub:time} Table~\ref{tab:filter_time} provides the average per-element reject time of all learned filters, taken across all the query sequences and space budgets that we have used in our experiments. They are expressed as percentage increment (or decrement) of the time required by the baseline. A first novel and interesting feature which emerges is that learned BF are sometimes faster than the baseline, which in principle is not expected, since learned variants have to query a classifier in addition to a classical BF. Our interpretation is that it can happen for two main reasons: 1) the adopted classifier is very fast and also effective, hence allowing in most cases to skip querying the backup filter; 2) the key set is very large, thus requiring a large baseline BF, whereas a good classifier can allow to sensibly drop the dimension of backup filters, thus making their invocation much faster. See for instance the case of DNA data, where most learned filters are faster that the baseline, with most classifiers.

Another intriguing behaviour concerns the reject time of ADA-BF, often the worst architecture in terms of this metric.
We believe it depends on the more complex procedure used in order to establish whether or  not to access the backup filter. Indeed, such a procedure is subject to tuning, which in turn can yield less or more complex  instances of the filter. As evident from our experiments, such a strategy does not always payoff.
We also observe sometimes a reversed order of the inference time of classifiers  (Table~\ref{tab:cl_time}) and the reject time of the corresponding filters. For instance, SVM is always the fastest classifier, but in some cases RF-based filters are faster (e.g., LBF and SLBF on Synthetic data). Even this behaviour is not so immediate to explain, and for sure it is related to the characteristics 1) and 2) mentioned above. Indeed, RF is the second fastest classifier, and it has an inference time just one order of magnitude slower that SVM, in average. On the other side, RF outperforms SVMs (even slightly), and this can reduce the number of queries to the backup filters and also their execution time (when the former is smaller). Finally, it is worth pointing out that the classifiers emerged as most effective in both classifier screening and learned BF analysis, namely NNs, are at least two orders of magnitude slower than the other classifiers, which must to be taken into account when configuring a learned BF, as we emphasize in the following discussion.


\section{Guidelines}\label{sec:filters_eval}
We summarize here our findings about the configuration of learned variants of Bloom Filters exploiting the prior knowledge of data complexity and available space budget.

\paragraph{Dataset Complexity and Classifier Choice.} We have roughly distinguished two main categories of data based on their complexity and the related behaviour of filters: easy dataset, having $\mathrm{F1v}\leq 0.35$, and hard dataset, having $\mathrm{F1v} > 0.35$.
The dataset complexity emerged as a central discriminant for the filter setup. Indeed,
when dealing with easy datasets,
the choice of the classifier becomes quite easy, and, independently of the space budget, it is always more convenient to select simple classifiers. Specifically, our experiments designate linear SVMs as best choice for the easiest datasets (those having almost linear separation boundary), and the smallest/simplest NNs for the more complex data in this category. In addition, the classifier inference time plays an important role for this data category: although a fastest classifier does not necessarily implies
a lower reject time of the corresponding filter (see Section~\ref{subsub:time}), when the average performance of two classifiers is close, then the inference time can be a discriminant feature for the classifier selection. But only in this case: see for instance the URL results, where the RF-$10$ performed just slightly better than SVMs, but although having an inference time one order of magnitude higher, the induced LBF has a lower reject time (see Section~\ref{subsub:time} for the relative discussion).
Surprisingly, this analysis has  been overlooked in the literature.
For instance, benchmark URL data falls in this category, but all previous experimental studies regarding learned BF on this data do not consider neither SVMs, nor NNs.

\begin{figure*}
\centering

\subfigure{
\includegraphics[width=0.25\textwidth]{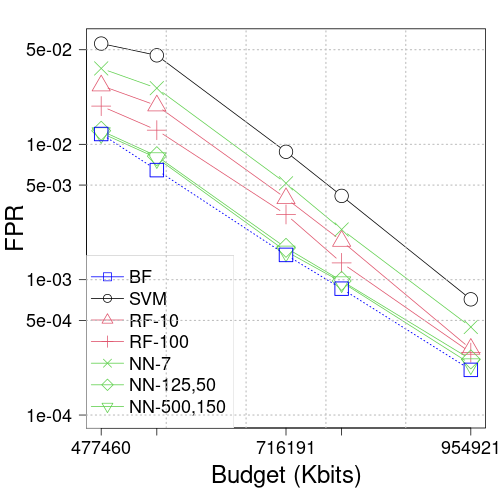}
}
\subfigure{
\includegraphics[width=0.25\textwidth]{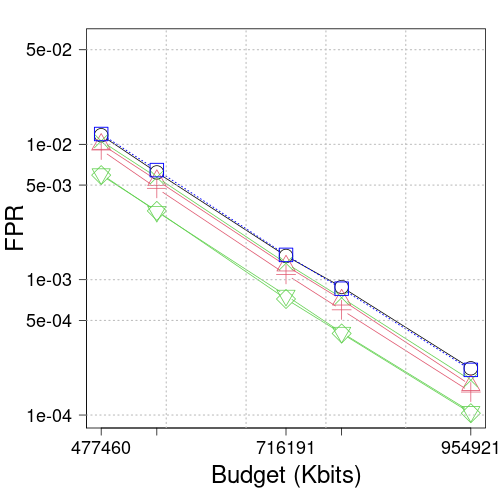}
}
\subfigure{
\includegraphics[width=0.25\textwidth]{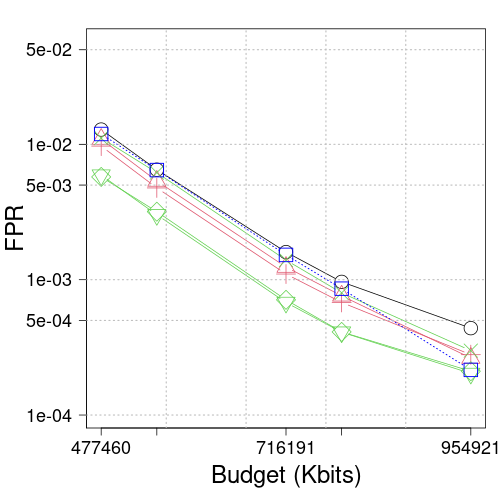}
}
\caption{Empirical false positive rate of LBF (left), SLFB (central), and ADA-BF (right) filters on DNA data. On the horizontal axis the different budgets configurations. Dotted blue line represents the baseline classical Bloom Filter. }\label{fig:filters_DNA}
\end{figure*}

For hard datasets instead, the space budget is central for the classifier choice.  Indeed, within the budget given by (\ref{eq:bloom}),
on synthetic datasets, having a relative small key set and accordingly a lower budget,
the choice is almost forced towards small although inaccurate classifiers, being the larger ones too demanding for the available budget. In particular, SVM is to be excluded due to the increased difficulty w.r.t.\ URL data, and for the remaining classifiers, we note that they behave very similarly (Figure~\ref{fig:results_synth}). Thus the most succinct ones are to be preferred in this case, namely the smallest NN.
As opposite, when the space budget increases, as it happens for DNA data, our findings suggest to learn more accurate classifiers, even if this requires the usage of a considerable budget fraction. Indeed, the gain induced by higher classification abilities in turn allows to save space when constructing the backup filter, to have consequently a smaller reject time, as well as an overall more efficient structure (cfr.\ Section~\ref{subsub:filtperf}).
This is also motivated by the fact that  to accurately train complex classifiers the sample size must be large enough~\cite{Raudys70}.
Therefore, on DNA data (Figures~\ref{fig:results_real}(b) and~\ref{fig:filters_DNA}), the most complex NNs resulted as the best choice.

\paragraph{Learned Bloom Filters Choice}
Regarding the choice of which learned BF is more suitable to use, our experiments reveal three main trends: 1) on benchmark data, that is those used also in the literature so far (URL data),
ADA-BF is confirmed as more effective variant in terms of FPR, when the budget is fixed;
2) however, its reject time is always and largely the highest one, thus suggesting to evaluate its usage in applications where fast responses are necessary (e.g., online applications). This subject includes also the classifier choice, since the most effective filters are in most cases induced by NNs, but they are also slower in terms of reject time of the counterparts induced by faster classifiers, and accordingly a trade-off FPR/reject time must be carefully investigated also when choosing the classifier; 3) SLBF is the filter most robust to data noise, and the only one able to benefit more even from classifier with poor performance (cfr.\ synthetic noisy and DNA data).
As a consequence, SBLF is clearly the filter to choose in presence of very complex data.
In particular, the point 3) is a new and quite unforeseen behaviour, emerged only thanks to the study of data complexity and relative noise injection procedure designed in this study, and which to some extent changes the ranking of most effective learned BF in practice,
since most real datasets are typically characterized by noise.

\begin{table}
\caption{Learned Bloom Filters average reject time, expressed as percentage of the baseline BF reject time. Positive (resp. negative) values indicate that the learned filter is slower (faster) than the baseline. Results are averaged across test queries and the filter space budgets considered. We remark that for DNA experiments another machine has been used w.r.t. synthetic and URL data (see Section~\ref{sec:HW}).}\label{tab:filter_time}
\centering
\small
\begin{tabular}{lrrr}
  \hline
\multicolumn{4}{c}{{\tt Synthetic Data} ($1.364\cdot 10^{-5}$)}\\[2pt]
\textbf{Classifier} &  \textbf{LBF} & \textbf{SLBF} & \textbf{ADA-BF}\\
SVM &  $18.4$ & $6.1$ & $151.2$\\
RF &  $-11.1$ & $-17.5$ & $112.8$\\
NN &  $106.9$ & $54.1$ & $159.3$\\
\multicolumn{4}{c}{{\tt URL Data} ($3.259\cdot 10^{-5}$)}\\[2pt]
SVM &  $22.6$ & $3.7$ & $3.9$\\
RF &  $6.6$ & $7.1$ & $9.7$\\
NN &  $43.9$ & $49.6$ & $35.3$\\
\multicolumn{4}{c}{{\tt DNA  Data ($4.817\cdot 10^{-5}$)}}\\[2pt]
SVM & $-12.5$ & $-11.7$ & $35.9$\\
RF-10 & $1.4$ & $-20.6$ & $32.0$\\
RF-100&  $19.8$ & $-7.4$ & $40.6$\\
NN-7  & $-5.0$ & $-12.0$ & $25.8$\\
NN-125,50&  $-3.6$ & $-7.5$ & $25.2$\\
NN-500,150& $-1.9$ & $-11.6$ & $39.2$\\
   \hline
\end{tabular}
\end{table}

\section{Conclusions and Future Developments}\label{sec:conclusions}
We have proposed an experimental methodology that can guide in the design and validation of learned Bloom Filters. The key point is to base the choice of the classifier to be used in a learned Bloom Filter on  the space budget of the entire data structure as well as the classification complexity of the input dataset. We have experimentally shown that our methodology yields useful indications, e.g.,  how robust are the current learned Boom Filters for the processing of datasets of increasing complexity. So far, only the ``easy to classify'' scenario has been considered in the Literature. A potential limitation of such results is that they might be dependent on the considered data; nonetheless, this is somehow inevitable due to the nature of Learned Data Structures.
Finally, we point out that the societal impacts of our contributions are in line with general-purpose Machine Learning technology.
Natural extensions of this research are as follows. As  already remarked, we have complied  with  an experimental setting coherent with the State of the Art. We can also consider the case in which the desired FPR is fixed  and one asks for the most succinct pair classifier-filter.
Moreover, in his seminal paper \cite{Mitz18}, Mitzenmacher has shown that Learned Bloom Filters can be quite sensitive to the input query distribution. Yet, no study is available to quantify this aspect. Our methodology can be extended also to those types   of analysis and work in this direction is in progress.










\section*{Acknowledgements}
This work has been supported by the Italian MUR PRIN project ``Multicriteria data structures and algorithms: from compressed to learned indexes, and beyond'' (Prot. 2017WR7SHH). Additional support to R.G. has been granted by Project INdAM - GNCS  ``Analysis and Processing of Big Data based on Graph Models''.

\clearpage

\bibliographystyle{plain}
\bibliography{biblio}

\begin{thebibliography}{10}

\bibitem{ali2006}
Shawkat Ali and Kate~A Smith.
\newblock On learning algorithm selection for classification.
\newblock {\em Applied Soft Computing}, 6(2):119--138, 2006.

\bibitem{amato2021lncs}
D.~Amato, G.~Lo Bosco, and R.~Giancarlo.
\newblock Learned sorted table search and static indexes in small model space
  ({E}xtended {A}bstract).
\newblock In {\em Proc. of the 20-th Italian Conference in Artificial
  Intelligence (AIxIA), to appear in Lecture Notes in Computer Science}, 2021.

\bibitem{Bloom70}
Burton~H. Bloom.
\newblock Space/time trade-offs in hash coding with allowable errors.
\newblock {\em Commun. ACM}, 13(7):422--426, July 1970.

\bibitem{Boffa21}
Antonio Boffa, Paolo Ferragina, and Giorgio Vinciguerra.
\newblock A ``learned'' approach to quicken and compress rank/select
  dictionaries.
\newblock In {\em Proceedings of the SIAM Symposium on Algorithm Engineering
  and Experiments (ALENEX)}, 2021.

\bibitem{RF}
L.~Breiman.
\newblock Random {F}orests.
\newblock {\em Machine Learning}, 45(1):5--32, 2001.

\bibitem{CART}
Leo {Breiman}, Jerome~H. {Friedman}, Richard~A. {Olshen}, and Charles~J.
  {Stone}.
\newblock {Classification and regression trees}.
\newblock {The Wadsworth Statistics/Probability Series. Belmont, California:
  Wadsworth International Group, a Division of Wadsworth, Inc. X, 358 p. {\$}
  29.25; {\$} 18.95 (1984).}, 1984.

\bibitem{Broder2005}
Andrei Broder and Michael Mitzenmacher.
\newblock {Network Applications of Bloom Filters: A Survey}.
\newblock In {\em Internet Mathematics}, volume~1, pages 636--646, 2002.

\bibitem{BRUZZONE1997}
L.~Bruzzone and S.B. Serpico.
\newblock Classification of imbalanced remote-sensing data by neural networks.
\newblock {\em Pattern Recognition Letters}, 18(11):1323--1328, 1997.

\bibitem{Cano13}
Jos\'{e}-Ram\'{o}n Cano.
\newblock Analysis of data complexity measures for classification.
\newblock {\em Expert Systems with Applications}, 40(12):4820--4831, 2013.

\bibitem{wegman79}
J.Lawrence Carter and Mark~N. Wegman.
\newblock Universal classes of hash functions.
\newblock {\em Journal of Computer and System Sciences}, 18(2):143--154, 1979.

\bibitem{GRU}
Kyunghyun Cho, Bart van Merri{\"e}nboer, Dzmitry Bahdanau, and Yoshua Bengio.
\newblock On the properties of neural machine translation: Encoder{--}decoder
  approaches.
\newblock In {\em Proceedings of {SSST}-8, Eighth Workshop on Syntax, Semantics
  and Structure in Statistical Translation}, pages 103--111, Doha, Qatar,
  October 2014. Association for Computational Linguistics.

\bibitem{chor2009genomic}
Benny Chor, David Horn, Nick Goldman, Yaron Levy, Tim Massingham, et~al.
\newblock Genomic {DNA} k-mer spectra: models and modalities.
\newblock {\em Genome Biol}, 10(10):R108, 2009.

\bibitem{LRegr}
David~R Cox.
\newblock The regression analysis of binary sequences.
\newblock {\em Journal of the Royal Statistical Society: Series B
  (Methodological)}, 20(2):215--232, 1958.

\bibitem{adabf}
Zhenwei Dai.
\newblock Adaptive learned bloom filter (ada-bf): Efficient utilization of the
  classifier.
\newblock \url{https://github.com/DAIZHENWEI/Ada-BF}, 2022.
\newblock Last checked on Nov.~8, 2022.

\bibitem{Dai}
Zhenwei Dai and Anshumali Shrivastava.
\newblock {A}daptive {L}earned {B}loom {F}ilter ({A}da-{BF}): Efficient
  utilization of the classifier with application to real-time information
  filtering on the web.
\newblock In {\em Advances in Neural Information Processing Systems},
  volume~33, pages 11700--11710. Curran Associates, Inc., 2020.

\bibitem{NaiveBC}
R.~O. Duda and P.~E. Hart.
\newblock {\em Pattern Classification and Scene Analysis}.
\newblock John Willey \& Sons, New York, 1973.

\bibitem{duda20}
Richard~O. Duda, Peter~E. Hart, and David~G. Stork.
\newblock {\em Pattern Classification, 2nd Edition}.
\newblock Wiley, 2000.

\bibitem{Elworth2020}
R.~A.~Leo Elworth, Qi~Wang, Pavan~K. Kota, C.~J. Barberan, Benjamin Coleman,
  Advait Balaji, Gaurav Gupta, Richard~G. Baraniuk, Anshumali Shrivastava, and
  Todd~J. Treangen.
\newblock To petabytes and beyond: recent advances in probabilistic and signal
  processing algorithms and their application to metagenomics.
\newblock {\em Nucleic acids research}, 48(10):5217--5234, Jun 2020.
\newblock 32338745[pmid], PMC7261164[pmcid], 5825624[PII].

\bibitem{FERRAGINA21}
Paolo Ferragina, Fabrizio Lillo, and Giorgio Vinciguerra.
\newblock On the performance of learned data structures.
\newblock {\em Theoretical Computer Science}, 871:107--120, 2021.

\bibitem{Ferragina:2020pgm}
Paolo Ferragina and Giorgio Vinciguerra.
\newblock The {PGM-index}: a fully-dynamic compressed learned index with
  provable worst-case bounds.
\newblock {\em {PVLDB}}, 13(8):1162--1175, 2020.

\bibitem{flores2014}
M~Julia Flores, Jos{\'e}~A G{\'a}mez, and Ana~M Mart{\'\i}nez.
\newblock Domains of competence of the semi-naive bayesian network classifiers.
\newblock {\em Information Sciences}, 260:120--148, 2014.

\bibitem{FreedmanStat}
David Freedman.
\newblock {\em Statistical {M}odels : {T}heory and {P}ractice}.
\newblock {Cambridge University Press}, 2005.

\bibitem{Fumagalli:2021}
G.~Fumagalli, D.~Raimondi, R.~Giancarlo, D.~Malchiodi, and M.~Frasca.
\newblock On the choice of general purpose classifiers in learned bloom
  filters: An initial analysis within basic filters.
\newblock In {\em Proceedings of the 11th International Conference on Pattern
  Recognition Applications and Methods (ICPRAM)}, pages 675--682, 2022.

\bibitem{haykin94}
Simon Haykin.
\newblock {\em Neural networks: a comprehensive foundation}.
\newblock Prentice Hall PTR, 1994.

\bibitem{He09}
Haibo He and E.A. Garcia.
\newblock Learning from imbalanced data.
\newblock {\em Knowledge and Data Engineering, IEEE Transactions on},
  21(9):1263--1284, Sept 2009.

\bibitem{Khalilia11}
Mohammed Khalilia, Sounak Chakraborty, and Mihail Popescu.
\newblock Predicting disease risks from highly imbalanced data using random
  forest.
\newblock {\em BMC Medical Informatics and Decision Making}, 11(1):51, Jul
  2011.

\bibitem{Kipf20}
Andreas Kipf, Ryan Marcus, Alexander van Renen, Mihail Stoian, Alfons Kemper,
  Tim Kraska, and Thomas Neumann.
\newblock Radixspline: A single-pass learned index.
\newblock In {\em Proc.\ of the Third International Workshop on Exploiting
  Artificial Intelligence Techniques for Data Management}, aiDM '20, pages
  1--5. Association for Computing Machinery, 2020.

\bibitem{btaa911}
Melanie Kirsche, Arun Das, and Michael~C Schatz.
\newblock {Sapling: accelerating suffix array queries with learned data
  models}.
\newblock {\em Bioinformatics}, 37(6):744--749, 10 2020.

\bibitem{Kirshe20}
Melanie Kirsche, Arun Das, and Michael~C Schatz.
\newblock {Sapling: accelerating suffix array queries with learned data
  models}.
\newblock {\em Bioinformatics}, 37(6):744--749, 10 2020.

\bibitem{Kraska21}
Tim Kraska.
\newblock Towards instance-optimized data systems.
\newblock {\em Proc. VLDB Endow.}, 14(12):3222–3232, oct 2021.

\bibitem{Kraska18}
Tim Kraska, Alex Beutel, Ed~H. Chi, Jeffrey Dean, and Neoklis Polyzotis.
\newblock The case for learned index structures.
\newblock In {\em Proc.\ of the 2018 Int.\ Conf.\ on Management of Data},
  SIGMOD '18, pages 489--504, New York, NY, USA, 2018. Association for
  Computing Machinery.

\bibitem{Liu22}
Qiyu Liu, Libin Zheng, Yanyan Shen, and Lei Chen.
\newblock Stable learned bloom filters for data streams.
\newblock {\em Proc. VLDB Endow.}, 13(12):2355–2367, sep 2020.

\bibitem{lorena2019}
Ana~C. Lorena, Lu\'{\i}s P.~F. Garcia, Jens Lehmann, Marcilio C.~P. Souto, and
  Tin~Kam Ho.
\newblock How complex is your classification problem? a survey on measuring
  classification complexity.
\newblock {\em ACM Comput. Surv.}, 52(5), sep 2019.

\bibitem{luengo2015}
Juli{\'a}n Luengo and Francisco Herrera.
\newblock An automatic extraction method of the domains of competence for
  learning classifiers using data complexity measures.
\newblock {\em Knowledge and Information Systems}, 42(1):147--180, 2015.

\bibitem{Mailtry21}
Marcel Maltry and Jens Dittrich.
\newblock A critical analysis of recursive model indexes.
\newblock {\em CoRR}, abs/2106.16166, 2021.

\bibitem{Marcus20}
Ryan Marcus, Andreas Kipf, Alexander van Renen, Mihail Stoian, Sanchit Misra,
  Alfons Kemper, Thomas Neumann, and Tim Kraska.
\newblock Benchmarking learned indexes.
\newblock {\em arXiv preprint arXiv:2006.12804}, 14:1--13, 2020.

\bibitem{Markus20b}
Ryan Marcus, Emily Zhang, and Tim Kraska.
\newblock {CDFS}hop: Exploring and optimizing learned index structures.
\newblock In {\em Proc.\ of the 2020 ACM SIGMOD Int.\ Conf.\ on Management of
  Data}, SIGMOD '20, pages 2789--2792, 2020.

\bibitem{neucom}
Giosuè~Cataldo Marinò, Alessandro Petrini, Dario Malchiodi, and Marco Frasca.
\newblock Deep neural networks compression: a comparative survey and choice
  recommendations.
\newblock {\em Neurocomputing}, 2022.

\bibitem{Mitz18}
Michael Mitzenmacher.
\newblock A model for learned bloom filters and optimizing by sandwiching.
\newblock In {\em Advances in Neural Information Processing Systems},
  volume~31. Curran Associates, Inc., 2018.

\bibitem{Mitz20}
Michael Mitzenmacher and Sergei Vassilvitskii.
\newblock Algorithms with predictions.
\newblock In Tim Roughgarden, editor, {\em Beyond the Worst-Case Analysis of
  Algorithms}, page 646–662. Cambridge University Press, 2021.

\bibitem{morik1999combining}
Katharina Morik, Peter Brockhausen, and Thorsten Joachims.
\newblock Combining statistical learning with a knowledge-based approach: a
  case study in intensive care monitoring.
\newblock Technical report, Technical Report, 1999.

\bibitem{Pickle}
{Python Software Foundation}.
\newblock pickle -- python object serialization.
\newblock \url{https://docs.python.org/3/library/pickle.html}, 2021.
\newblock Last checked on May.~17, 2022.

\bibitem{Rahman21}
Amatur Rahman and Paul Medevedev.
\newblock Representation of k-mer sets using spectrum-preserving string sets.
\newblock {\em Journal of Computational Biology}, 28(4):381--394, 2021.
\newblock PMID: 33290137.

\bibitem{supp-software}
Davide Raimondi and Giacomo Fumagalli.
\newblock A critical analysis of classifier selection in learned bloom filters
  -- supporting software.
\newblock \url{https://github.com/RaimondiD/LBF_ADABF_experiment}, 2022.
\newblock Last checked on Nov.~8, 2022.

\bibitem{Raudys70}
S~Raudys.
\newblock On the problems of sample size in pattern recognition.
\newblock In {\em Detection, pattern recognition and experiment design: Vol. 2.
  Proceedings of the 2nd all-union conference statistical methods in control
  theory}. Publ. House" Nauka", 1970.

\bibitem{Solomo}
Brad Solomon and Carl Kingsford.
\newblock Fast search of thousands of short-read sequencing experiments.
\newblock {\em Nature Biotechnology}, 34(3):300--302, Mar 2016.

\bibitem{Kraskap}
Kapil Vaidya, Eric Knorr, Tim Kraska, and Michael Mitzenmacher.
\newblock Partitioned learned bloom filters.
\newblock In {\em International Conference on Learning Representations}, 2021.

\bibitem{VanHulse07}
Jason Van~Hulse, Taghi~M. Khoshgoftaar, and Amri Napolitano.
\newblock Experimental perspectives on learning from imbalanced data.
\newblock In {\em Proceedings of the 24th International Conference on Machine
  Learning}, ICML '07, pages 935--942, New York, NY, USA, 2007. ACM.

\bibitem{wegman81}
Mark~N. Wegman and J.Lawrence Carter.
\newblock New hash functions and their use in authentication and set equality.
\newblock {\em Journal of Computer and System Sciences}, 22(3):265--279, 1981.

\bibitem{wu21}
Qingtao Wu, Qianyu Wang, Mingchuan Zhang, Ruijuan Zheng, Junlong Zhu, and
  Jiankun Hu.
\newblock Learned bloom-filter for the efficient name lookup in
  information-centric networking.
\newblock {\em Journal of Network and Computer Applications}, 186:103077, 04
  2021.

\bibitem{FFNN}
Andreas Zell.
\newblock {\em Simulation neuronaler Netze}.
\newblock habilitation, Uni Stuttgart, 1994.

\end{thebibliography}

\end{document}